\documentclass{article}
\usepackage{graphicx}
\usepackage{bm}
\usepackage{amsmath}
\usepackage{amssymb}
\usepackage{amsfonts}
\usepackage{natbib}

\begin{document}

\title{Multi-view Anomaly Detection\\via Probabilistic Latent Variable Models
}


\author{Tomoharu Iwata         
\thanks{NTT Communication Science Laboratories}
\and
        Makoto Yamada 
\thanks{Yahoo Labs}
}
\date{}

\maketitle

\begin{abstract}
We propose a nonparametric Bayesian probabilistic latent variable model for multi-view anomaly detection, which is the task of finding instances that have inconsistent views. With the proposed model, all views of a non-anomalous instance are assumed to be generated from a single latent vector. On the other hand, an anomalous instance is assumed to have multiple latent vectors, and its different views are generated from different latent vectors. By inferring the number of latent vectors used for each instance with Dirichlet process priors, we obtain multi-view anomaly scores. The proposed model can be seen as a robust extension of probabilistic canonical correlation analysis for noisy multi-view data. We present Bayesian inference procedures for the proposed model based on a stochastic EM algorithm. The effectiveness of the proposed model is demonstrated in terms of performance when detecting multi-view anomalies and imputing missing values in multi-view data with anomalies.
\end{abstract}

\section{Introduction}

There has been great interest in multi-view learning,
in which data are obtained from various information sources~\citep{blum1998combining,bickel2004multi,sindhwani2008rkhs}.
In a wide variety of applications, data are naturally 
comprised of multiple views.
For example, 
an image can be represented by color, texture and shape information;
a web page can be represented by words, images and URLs occurring on 
in the page;
and a video can be represented by audio and visual features.
In this paper, we consider the task of finding anomalies 
in multi-view data.
The task is called horizontal anomaly detection~\citep{gao2011spectral},
or multi-view anomaly detection~\citep{liu2012using}.
Anomalies in multi-view data are instances that have inconsistent 
features across multiple views.
Multi-view anomaly detection is different 
from standard (single-view) anomaly detection.
Single-view anomaly detection finds instances
that do not conform to expected behavior~\citep{chandola2009anomaly}.

\begin{figure}[t!]
\centering
(c) Latent space\\
\includegraphics[width=15em]{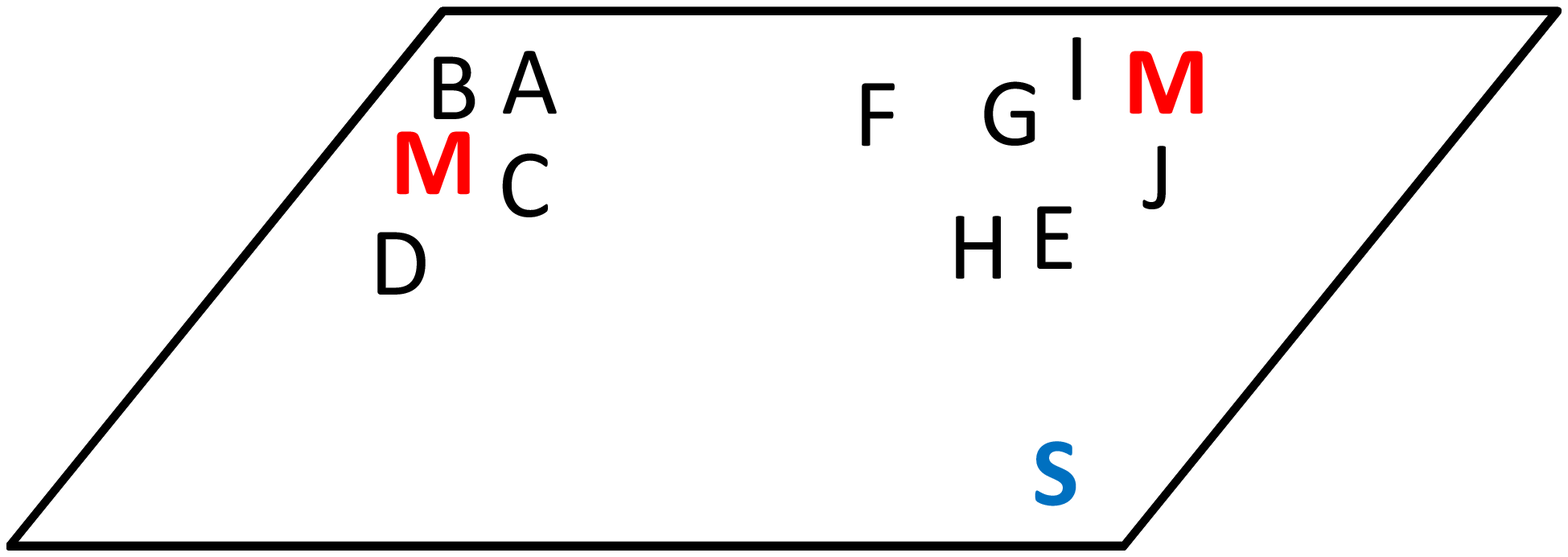}
\\
\vspace{1em}
\begin{tabular}{cc}
$\bm{W}_{1}$ {\Large $\swarrow$} & {\Large $\searrow$} $\bm{W}_{2}$
\vspace{0.5em}
\\
\includegraphics[width=9em]{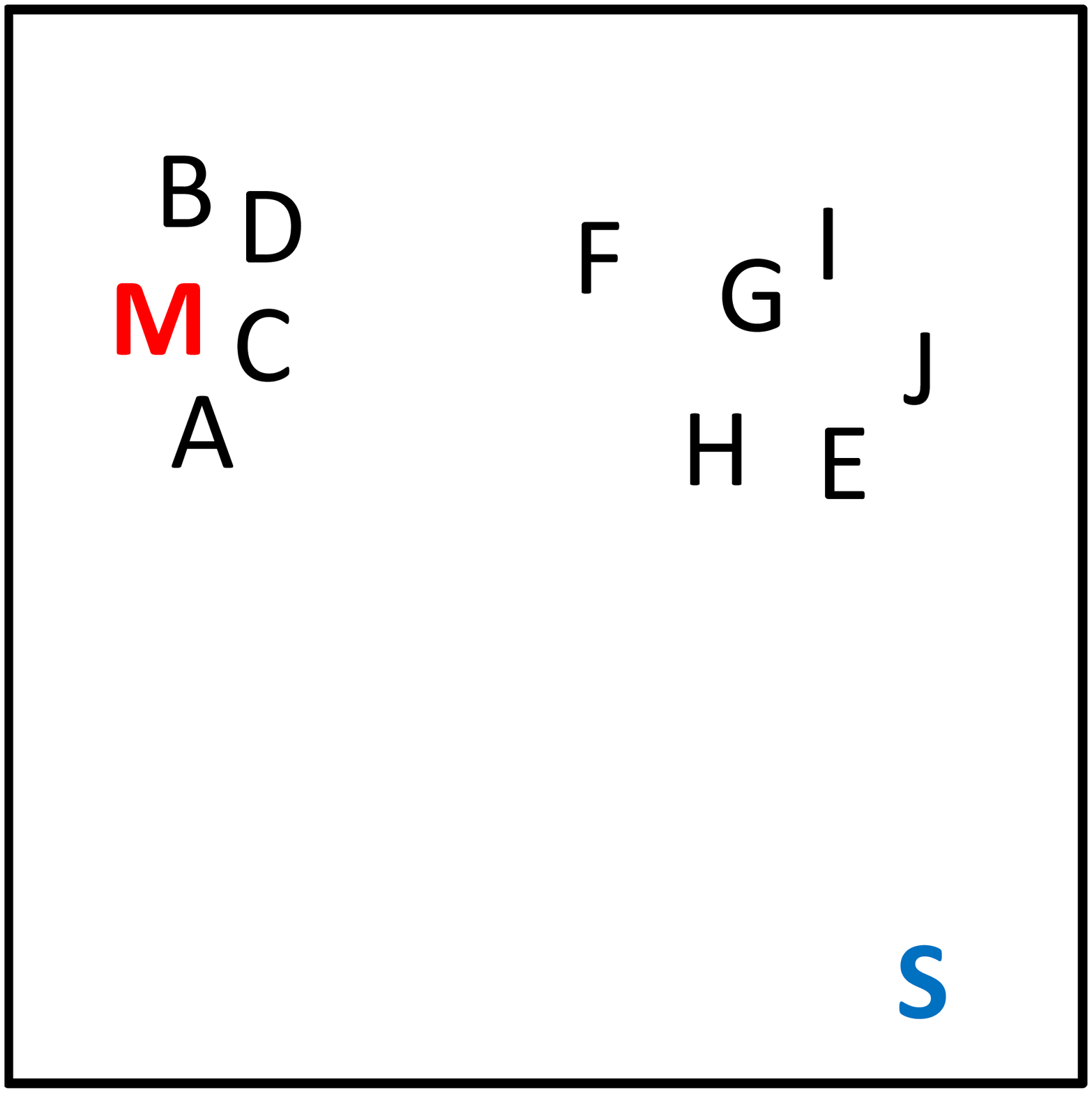}
&
\includegraphics[width=9em]{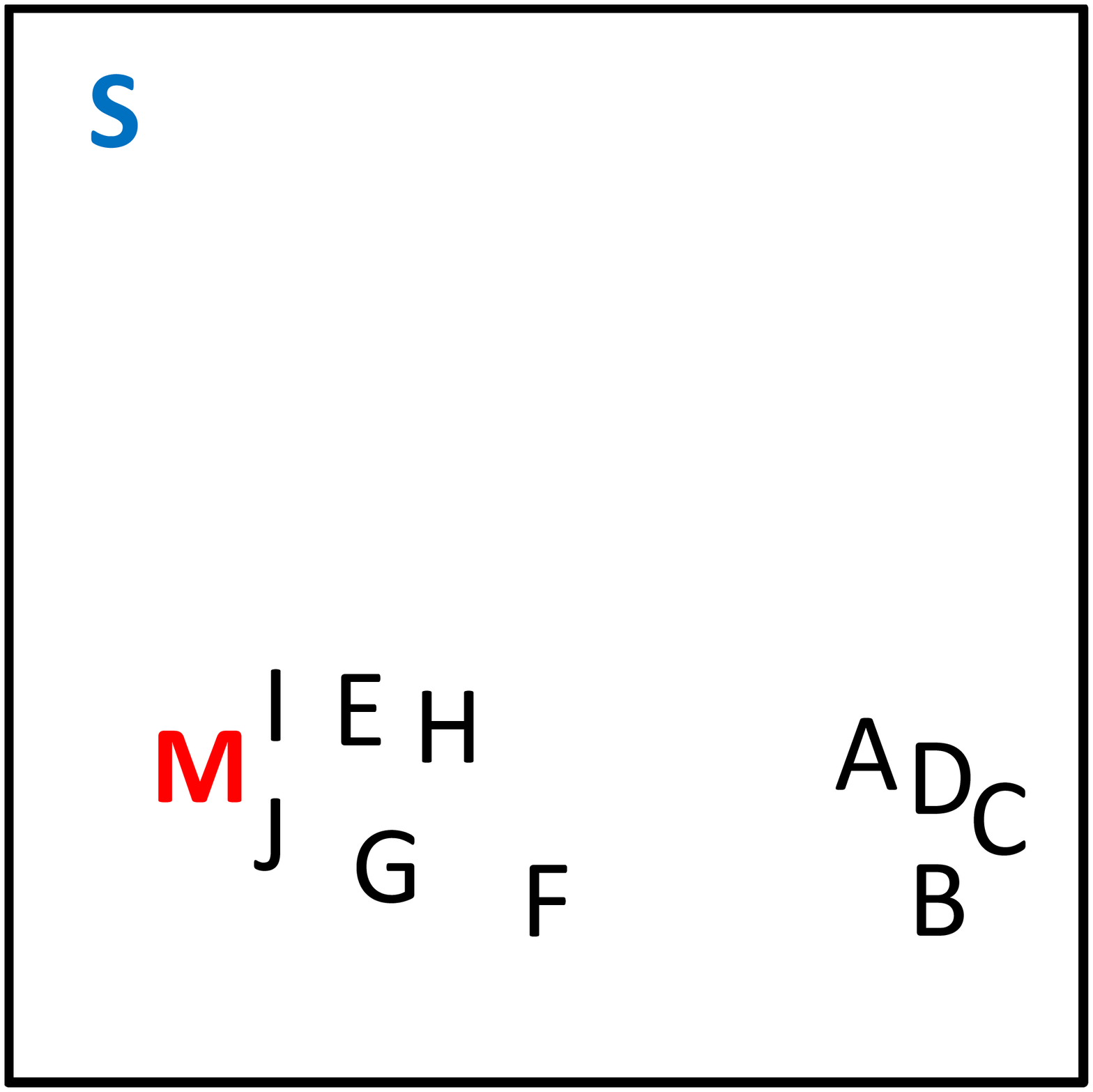}
\\
(a) Observed view 1 & (b) Observed view 2
\end{tabular}
\caption{A multi-view anomaly `M' and a single-view anomaly `S' in a two-view data set. Each letter represents an instance, and the same letter indicates the same instance. $\bm{W}_{d}$ is a projection matrix for view $d$.}
\label{fig:multiview_anomaly}
\end{figure}

Figure~\ref{fig:multiview_anomaly} (a,b) shows the difference between
a multi-view anomaly and a single-view anomaly in a two-view data set.
`M' is a multi-view anomaly 
since `M' belongs to different clusters in different views 
(`A--D' cluster in View 1 and `E--J' cluster in View 2) 
and views of `M' are not consistent. 
`S' is a single-view anomaly 
since `S' is located far from other instances in each view.
However, both views of `S' have the same relationship with the others 
(they are far from the other instances), 
and then `S' is {\it not} a multi-view anomaly.
Single-view anomaly detection methods, 
such as one-class support vector machines~\citep{scholkopf2001estimating},
consider that `S' is anomalous.
On the other hand, 
we would like to develop a multi-view anomaly detection method
that detects `M' as anomaly, but not `S'.

Multi-view anomaly detection can be used for many applications,
such as information disparity management~\citep{duh2013managing},
purchase behavior analysis~\citep{gao2011spectral},
malicious insider detection~\citep{liu2012using}, and
user aggregation from multiple databases.
In information disparity management, 
multiple views can be obtained from documents 
written in different languages such as Wikipedia.
Multi-view anomaly detection tries to find documents 
that contain different information across different languages.
In purchase behavior analysis,
multiple views for each item can be defined
as its genre and its purchase history, 
i.e. a set of users who purchased the item.
Multi-view anomaly detection can find movies inconsistently
purchased by users based on the movie genre.

We propose a probabilistic latent variable model 
for multi-view anomaly detection.
With the proposed model, there is a latent space 
that is shared across all views.
We assume that all views 
of a non-anomalous (normal) instance 
are generated using a single latent vector.
On the other hand,
an anomalous instance is assumed to have multiple latent vectors,
and its different views are generated using different latent vectors,
which indicates inconsistency across different views of the instance.
Figure~\ref{fig:multiview_anomaly} (c) shows an example of a latent space
shared by the two-view data. 
Two views of every non multi-view anomaly 
can be generated from a latent vector using
view-dependent projection matrices.
On the other hand, since two views of multi-view anomaly 
`M' are not consistent, two latent vectors are required 
to generate the two views using the projection matrices.

Since the number of latent vectors for each instance is unknown,
we automatically infer it from the given data by using 
Dirichlet process priors.
The inference of the proposed model is based on a stochastic EM algorithm.
In the E-step, a latent vector is assigned for each view of each instance
using collapsed Gibbs sampling while analytically 
integrating out latent vectors.
In the M-step, projection matrices for mapping latent vectors 
into observations are estimated by maximizing the joint likelihood.
By alternately iterating E- and M-steps, we infer 
the number of latent vectors used in each instance
and calculate its anomaly score from the probability of using 
more than one latent vector.

The proposed model can be seen 
as a robust extension of probabilistic canonical correlation analysis 
(PCCA)~\citep{bach2005,wang2007}.
PCCA assumes that each instance has a latent vector,
and the estimated parameters suffer effects from anomalies.
In contrast, by preparing multiple latent vectors for anomalies,
the proposed model can infer latent vectors and projection matrices
properly without influence of multi-view anomalies.
 

\section{Related Work}
\label{sec:related}

Anomaly detection has had a wide variety of applications,
such as
credit card fraud detection~\citep{aleskerov1997cardwatch},
intrusion detection for network security~\citep{portnoy2001intrusion},
and analysis for healthcare data~\citep{antonie2001application}.
However, most existing anomaly detection techniques assume data 
with a single view, i.e. a single observation feature set.

A number of anomaly detection methods for two-view data have been 
proposed~\citep{gao2010community,shekhar2002detecting,song2007conditional,sun2005neighborhood,wang2009discovering}.
However, they cannot be used for data with more than two views.
\citet{gao2011spectral} proposed 
a HOrizontal Anomaly Detection algorithm (HOAD)
for finding anomalies from multi-view data.
With HOAD, all views of all instances are simultaneously embedded
in a low-dimensional space 
with the constraints that different views of the same instance are embedded 
close together using a spectral framework~\citep{ng2002spectral}.
Then, anomalies are detected according to distance between 
the embedded locations of different views of each instance.
In HOAD, there are hyperparameters including a weight for the constraint
that require the data to be labeled as anomalous or not for tuning,
and the performance is sensitive to the hyperparameters.
On the other hand, the parameters with the proposed model 
can be estimated from the given multi-view data 
without label information by maximizing the likelihood.
In addition, because the proposed model is a probabilistic generative model,
we can extend it in a probabilistically principled manner, for example,
for handling missing data and combining with other probabilistic models.
\citet{liu2012using} proposed multi-view anomaly detection methods 
using consensus clustering. They found anomalies based on the inconsistency
of clustering results across multiple views. 
Therefore, they cannot find inconsistency within a cluster.
\citet{christoudiasmulti} proposed a method for filtering
instances that are corrupted by background noise from multi-view data.
The multi-view anomalies considered in this paper include
not only instances corrupted by background noise but also
instances categorized into different foreground classes across views,
and instances with inconsistent views even if they belong to the same cluster.

The proposed model is related to 
probabilistic canonical correlation analysis (PCCA).
When every instance has only one latent vector, 
the proposed model can be seen as PCCA.
PCCA, or canonical correlation analysis (CCA), 
can be used for multi-view anomaly detection.
With PCCA, a latent vector that is shared by all views for each instance 
and a linear projection matrix for each view are estimated
by maximizing the likelihood, 
or minimizing the reconstruction error of the given data.
The reconstruction error for each instance
can be used as an anomaly score.
However, the reconstruction errors are not reliable
because they are calculated from parameters 
that are estimated using data with anomalies 
by assuming that all of the instances are non-anomalous.
On the other hand, because the proposed model simultaneously estimates 
the parameters and infers anomalies,
the estimated parameters are not contaminated by the anomalies.
With some CCA-related methods, each latent vector is factorized into
shared and private components across 
different views~\citep{archambeau2009,jia2010factorized,salzmann2010factorized,virtanen2011bayesian}.
They assume that every instance has shared and private parts that
are the same dimensionality for all instances,
and they cannot be used for multi-view anomaly detection.
In contrast, the proposed model can detect anomalies by assuming that
non-anomalous instances have only shared latent vectors,
and anomalies have private latent vectors.
Recently, \citet{alvarez2013} proposed 
a multi-view anomaly detection method.
However, since the method is based on clustering,
it cannot find anomalies when there are no clusters in the given data.



\section{Proposed Model}
\label{sec:model}


Suppose that we are given $N$ instances with $D$ views
$\bm{X}=\{\bm{X}_{n}\}_{n=1}^{N}$,
where $\bm{X}_{n}=\{\bm{x}_{nd}\}_{d=1}^{D}$ is a set of 
multi-view observation vectors for the $n$th instance,
and $\bm{x}_{nd}\in \mathbb{R}^{M_{d}}$ 
is the observation vector of the $d$th view.
The task is to find anomalous instances
that have inconsistent observation features across multiple views.

We propose a probabilistic latent variable model for this task.
The proposed model assumes that
each instance has
potentially a countably infinite number of latent vectors
$\bm{Z}_{n}=\{\bm{z}_{nj}\}_{j=1}^{\infty}$,
where $\bm{z}_{nj}\in\mathbb{R}^{K}$.
Each view of an instance $\bm{x}_{nd}$ is generated
depending on a view-specific projection matrix 
$\bm{W}_{d}\in\mathbb{R}^{M_{d}\times K}$
and a latent vector $\bm{z}_{ns_{nd}}$that is selected from 
a set of latent vectors $\bm{Z}_{n}$.
Here, $s_{nd}\in\{1,\cdots,\infty\}$ is the latent vector assignment 
of $\bm{x}_{nd}$.
When the instance is non-anomalous and all its views are consistent,
all of the views are generated from
a single latent vector.
In other words, the latent vector assignments for all views
are the same, $s_{n1}=s_{n2}=\cdots=s_{nD}$.
When it is an anomaly and some views are inconsistent,
different views are generated from different latent vectors,
and some latent vector assignments are different,
i.e. $s_{nd}\neq s_{nd'}$ for some $d\neq d'$.

Specifically, 
the proposed model is an infinite mixture model,
where the probability for the $d$th view of the $n$th instance is given by
\begin{align}
p(\bm{x}_{nd}|\bm{Z}_{n},\bm{W}_{d},\bm{\theta}_{n},\alpha)
=\sum_{j=1}^{\infty}\theta_{nj}{\cal N}(\bm{x}_{nd}|\bm{W}_{d}\bm{z}_{nj},\alpha^{-1}\bm{I}),
\end{align}
where 
$\bm{\theta}_{n}=\{\theta_{nj}\}_{j=1}^{\infty}$ are the mixture weights, 
$\theta_{nj}$ represents the probability of choosing the $j$th latent vector,
$\alpha$ is a precision parameter,
${\cal N}(\bm{\mu},\bm{\Sigma})$ denotes the Gaussian distribution 
with mean $\bm{\mu}$ and covariance matrix $\bm{\Sigma}$,
and $\bm{I}$ is the identity matrix.
We use a Dirichlet process for the prior of mixture weight $\bm{\theta}_{n}$.
Its use enables us to automatically infer the number of latent vectors 
for each instance from the given data.

\begin{figure}[t!]
\centering
\includegraphics[height=9em]{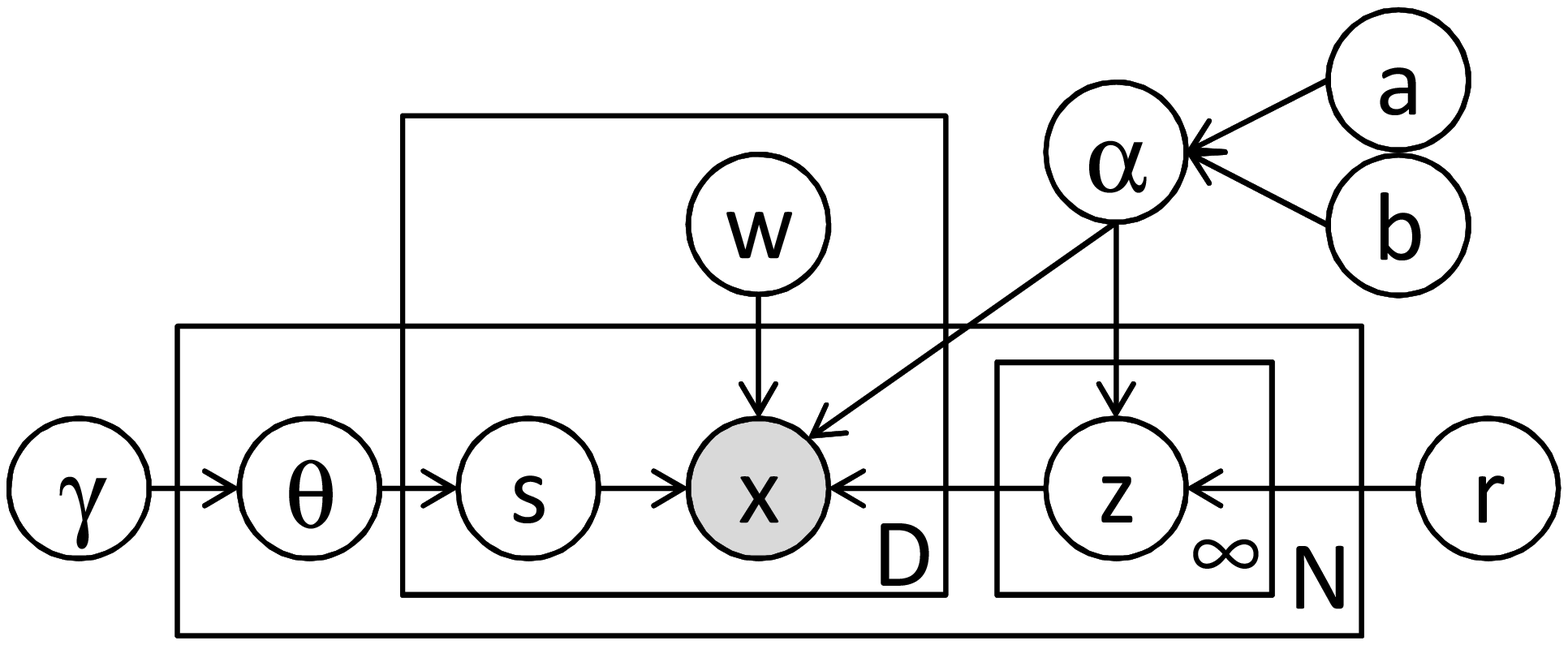}
\caption{Graphical model representation of the proposed model.}
\label{fig:graphical}
\end{figure}

The complete generative process of the proposed model 
for multi-view instances $\bm{X}$ is as follows,
\begin{enumerate}
\item Draw a precision parameter\\
$\alpha \sim \text{Gamma}(a,b)$
\item For each instance: $n=1,\dots,N$
\begin{enumerate}
\item Draw mixture weights\\
$\bm{\theta}_{n} \sim \text{Stick}(\gamma)$
\item For each latent vector: $j=1,\dots,\infty$
\begin{enumerate}
\item Draw a latent vector\\
$\bm{z}_{nj} \sim {\cal N}(\bm{0},(\alpha r)^{-1}\bm{I})$
\end{enumerate}
\item For each view: $d=1,\dots,D$
\begin{enumerate}
\item Draw a latent vector assignment\\
$s_{nd}\sim \text{Discrete}(\bm{\theta}_{n})$
\item Draw an observation vector\\
$\bm{x}_{nd}\sim {\cal N}(\bm{W}_{d}\bm{z}_{ns_{nd}},\alpha^{-1}\bm{I})$
\end{enumerate}
\end{enumerate}
\end{enumerate}
Here, $\text{Stick}(\gamma)$ is the stick-breaking process~\citep{sethuraman94}
that generates mixture weights for a Dirichlet process 
with concentration parameter $\gamma$,
and
$r$ is the relative precision for latent vectors.
$\alpha$ is shared for observation and latent vector precision because it makes it possible to analytically integrate out $\alpha$ as shown in (\ref{eq:likelihood}).
Figure~\ref{fig:graphical} shows a graphical model representation of
the proposed model,
where the shaded and unshaded nodes 
indicate observed and latent variables, respectively.

The joint probability of the data $\bm{X}$ and 
the latent vector assignments $\bm{S}=\{\{s_{nd}\}_{d=1}^{D}\}_{n=1}^{N}$
is given by
\begin{align}
p(\bm{X},\bm{S}|\bm{W},a,b,r,\gamma)
=p(\bm{S}|\gamma)p(\bm{X}|\bm{S},\bm{W},a,b,r),
\label{eq:joint}
\end{align}
where $\bm{W}=\{\bm{W}_{d}\}_{d=1}^{D}$.
Because we use conjugate priors, we can analytically 
integrate out mixture weights
$\bm{\Theta}=\{\bm{\theta}_{n}\}_{n=1}^{N}$,
latent vectors $\bm{Z}$, and precision parameter $\alpha$.
Here, we use a Dirichlet process prior 
for multinomial parameter $\bm{\theta}_{n}$,
and a Gaussian-Gamma prior for latent vector $\bm{z}_{nj}$.
By integrating out mixture weights $\bm{\Theta}$,
the first factor is calculated by
\begin{align} 
p(\bm{S}|\gamma)=\prod_{n=1}^{N}\frac{\gamma^{J_{n}}\prod_{j=1}^{J_{n}}(N_{nj}-1)!}{\gamma(\gamma+1)\cdots(\gamma+D-1)},
\label{eq:dp}
\end{align}
where $N_{nj}$ represents the number of views assigned 
to the $j$th latent vector in the $n$th instance,
and $J_{n}$ is the number of latent vectors of the $n$th instance
for which $N_{nj}>0$.
By integrating out latent vectors $\bm{Z}$ and precision parameter $\alpha$,
the second factor of (\ref{eq:joint}) is calculated by 
\begin{align}
p(\bm{X}|\bm{S},\bm{W},a,b,r)
=
(2\pi)^{-\frac{N\sum_{d}M_{d}}{2}}
r^{\frac{K\sum_{n}J_{n}}{2}}
\frac{b^{a}}{b'^{a'}}
\frac{\Gamma(a')}{\Gamma(a)}
\prod_{n=1}^{N}\prod_{j=1}^{J_{n}}|\bm{C}_{nj}|^{\frac{1}{2}},
\label{eq:likelihood}
\end{align}
where
\begin{align}
a'=a+\frac{N\sum_{d=1}^{D}M_{d}}{2},
\end{align}
\begin{align}
b'=b+\frac{1}{2}\sum_{n=1}^{N}\sum_{d=1}^{D}\bm{x}_{nd}^{\top}\bm{x}_{nd}
-\frac{1}{2}\sum_{n=1}^{N}\sum_{j=1}^{J_{n}}\bm{\mu}_{nj}^{\top}\bm{C}_{nj}^{-1}\bm{\mu}_{nj},
\end{align}
\begin{align}
\bm{\mu}_{nj} = \bm{C}_{nj}\sum_{d:s_{nd}=j}\bm{W}_{d}^{\top}\bm{x}_{nd},
\end{align}
\begin{align}
\bm{C}_{nj}^{-1}=\sum_{d:s_{nd}=j}\bm{W}_{d}^{\top}\bm{W}_{d}+r\bm{I}.
\end{align}
The posterior for the precision parameter $\alpha$
is given by 
\begin{align}
p(\alpha|\bm{X},\bm{S},\bm{W},a,b)=\text{Gamma}(a',b'),
\end{align}
and the posterior for the latent vector $\bm{z}_{nj}$
is given by
\begin{align}
p(\bm{z}_{nj}|\bm{X},\bm{S},\bm{W},r)=\mathcal{N}(\bm{\mu}_{nj},\alpha^{-1}\bm{C}_{nj}).
\end{align}

The proposed model is a generalization of either 
probabilistic principal component analysis 
(PPCA)~\citep{tipping1999probabilistic} or
probabilistic canonical correlation analysis (PCCA).
When all views are generated from different latent vectors 
for every instance,
the proposed model corresponds to PPCA 
that is performed independently for each view.
When all views are generated from a single latent vector
for every instance,
the proposed model corresponds to PCCA with spherical noise.

\section{Inference}
\label{sec:inference}

We describe inference procedures for the proposed model 
based on a stochastic EM algorithm,
in which collapsed Gibbs sampling of latent vector assignments $\bm{S}$
and the maximum joint likelihood estimation of 
projection matrices $\bm{W}$
are alternately iterated
while analytically integrating out the latent vectors $\bm{Z}$,
mixture weights $\bm{\Theta}$
and precision parameter $\alpha$.
By integrating out latent vectors,
we do not need to explicitly infer the latent vectors,
leading to a robust and fast-mixing inference.

Let $\ell=(n,d)$ 
be the index of the $d$th view of the $n$th instance
for notational convenience.
In the E-step, 
given the current state of all but one latent assignment $s_{\ell}$,
a new value for $s_{\ell}$ is sampled from 
$\{1,\cdots,J_{n\setminus \ell}+1\}$
according to the following probability,
\begin{align}
p(s_{\ell}=j|\bm{X},\bm{S}_{\setminus \ell},\bm{W},a,b,r,\gamma)
\propto 
\frac{p(s_{\ell}=j,\bm{S}_{\setminus \ell}|\gamma)}
{p(\bm{S}_{\setminus \ell}|\gamma)}
\cdot
\frac{p(\bm{X}|s_{\ell}=j,\bm{S}_{\setminus \ell},\bm{W},a,b,r)}
{p(\bm{X}_{\setminus \ell}|\bm{S}_{\setminus \ell},\bm{W},a,b,r)},
\label{eq:estep}
\end{align}
where $\setminus \ell$ represents a value or set excluding the $d$th view
of the $n$th instance.
 
The first factor is given by
\begin{align}
\frac{p(s_{\ell}=j,\bm{S}_{\setminus \ell}|\gamma)}
{p(\bm{S}_{\setminus \ell}|\gamma)}
=
\left\{
\begin{array}{ll}
\frac{N_{nj\setminus \ell}}{D-1+\gamma} & \text{if $j\leq J_{n\setminus \ell}$}\\
\frac{\gamma}{D-1+\gamma} & \text{if $j=J_{n\setminus \ell}+1$},\\
\end{array}
\right.
\end{align}
using (\ref{eq:dp}), where 
$j\leq J_{n\setminus \ell}$ is for existing latent vectors,
and $j=J_{n\setminus \ell}+1$ is for a new latent vector.
By using (\ref{eq:likelihood}),
the second factor is given by
\begin{align}
\lefteqn{\frac{p(\bm{X}|s_{\ell}=j,\bm{S}_{\setminus \ell},\bm{W},a,b,r)}
{p(\bm{X}_{\setminus \ell}|\bm{S}_{\setminus \ell},\bm{W},a,b,r)}}
\nonumber\\
&=(2\pi)^{-\frac{M_{d}}{2}}
r^{I(j=J_{n\setminus \ell}+1)\frac{K}{2}}
\frac{b_{\setminus \ell}'^{a_{\setminus \ell}'}}
{b_{s_{\ell}=j}'^{a_{s_{\ell}=j}^{\prime}}}
\frac{\Gamma(a_{s_{\ell}=j}')}{\Gamma(a_{\setminus \ell}')}
\frac{|\bm{C}_{j,s_{\ell}=j}|^\frac{1}{2}}{|\bm{C}_{j\setminus \ell}|^\frac{1}{2}},
\end{align}
where 
$I(\cdot)$ represents the indicator function,
i.e. $I(A)=1$ if $A$ is true and $0$ otherwise,
and 
subscript $s_{\ell}=j$ indicates the value 
when $\bm{x}_{\ell}$ is assigned to the $j$th latent vector as follows,
\begin{align}
a_{s_{\ell}=j}^{\prime}=a^{\prime}, 
\end{align}
\begin{align}
b_{s_{\ell}=j}^{\prime}=b_{\setminus \ell}'+\frac{1}{2}\bm{x}_{\ell}^{\top}\bm{x}_{\ell}+\frac{1}{2}\bm{\mu}_{nj\setminus \ell}^{\top}\bm{C}_{nj\setminus \ell}^{-1}\bm{\mu}_{nj\setminus \ell}
-\frac{1}{2}\bm{\mu}_{nj,s_{\ell}=j}^{\top}\bm{C}_{nj,s_{\ell}=j}^{-1}\bm{\mu}_{nj,s_{\ell}=j},
\end{align}
\begin{align}
\bm{\mu}_{nj,s_{\ell}=j}=\bm{C}_{nj,s_{\ell}=j}(\bm{W}_{d}^{\top}\bm{x}_{\ell}+\bm{C}_{nj\setminus \ell}^{-1}\bm{\mu}_{nj\setminus \ell}),
\end{align}
\begin{align}
 \bm{C}_{nj,s_{\ell}=j}^{-1}=\bm{W}_{d}^{\top}\bm{W}_{d}+\bm{C}_{nj\setminus \ell}^{-1}.
\end{align}
Intuitively, if the current view cannot be modeled well 
by existing latent vectors, a new latent vector is used, 
which indicates that the view is inconsistent with the other views.

In the M-step, 
the projection matrices $\bm{W}$
are estimated by maximizing the logarithm 
of the joint likelihood (\ref{eq:joint}).
We can maximize the likelihood
by using a gradient-based numerical optimization method
such as the quasi-Newton method~\citep{nocedal1980updating}.
The gradient of the joint log likelihood is calculated by
\begin{align}
\lefteqn{\frac{\partial \log p(\bm{X},\bm{S}|\bm{W},a,b,r,\gamma)}{\partial \bm{W}_{d}}}
\nonumber\\
&=-\bm{W}_{d}\sum_{n=1}^{N}\sum_{j=1}^{J_{n}}\bm{C}_{nj}
-\frac{a'}{b'}
\sum_{n=1}^{N}
\Bigl(
\bm{W}_{d}\bm{\mu}_{ns_{nd}}\bm{\mu}_{ns_{nd}}^{\top}
-\bm{x}_{nd}\bm{\mu}_{ns_{nd}}^{\top}
\Bigr).
\label{eq:mstep}
\end{align}
When we iterate the E-step that samples the latent vector assignment $s_{nd}$
by employing (\ref{eq:estep}) for each view
$d=1,\dots,D$ in each instance $n=1,\dots,N$,
and the M-step that maximizes the joint likelihood using (\ref{eq:mstep}) 
with respect to the projection matrix $\bm{W}_{d}$ 
for each view $d=1,\dots,D$,
we obtain an estimate of the latent vector assignments and 
projection matrices.

For an anomaly score, 
we use the probability that the instance uses more than one latent vector.
It is estimated by using the samples obtained in the inference as follows,
\begin{align}
v_{n} = \frac{1}{H}\sum_{h=1}^{H}I(J_{n}^{(h)}>1),
\end{align}
where $J_{n}^{(h)}$ is the number of latent vectors 
used by the $n$th instance
in the $h$th iteration of the Gibbs sampling after the burn-in period,
and $H$ is the number of the iterations.

We can use cross-validation to select an appropriate dimensionality for the latent space $K$. With cross-validation, we assume that some features are missing from the given data, and infer the model with a different $K$. Then, we select the smallest $K$ value that has performed the best at predicting missing values.

\section{Experiments}
\label{sec:experiments}

\subsection{Data}
We evaluated the proposed model quantitatively by using 11 data sets,
which we obtained from the LIBSVM data sets~\citep{chang2011libsvm}.
We generated multiple views by randomly splitting the features,
and anomalies were added by swapping views of two randomly selected instances
as \citet{gao2011spectral} did in their experiments.
Splitting data does not generate anomalies. 
Therefore, we can evaluate methods while controlling the anomaly rate properly. 
By swapping, although single-view anomalies cannot be created 
since the distribution for each view does not change, 
multi-view anomalies are created.

\subsection{Comparing methods}
We compared the proposed model with 
probabilistic canonical correlation analysis (PCCA),
horizontal anomaly detection (HOAD)~\citep{gao2011spectral},
consensus clustering based anomaly detection (CC)~\citep{liu2012using},
and one-class support vector machine (OCSVM)~\citep{scholkopf2001estimating}.
For PCCA, we used the proposed model 
in which the number of latent vectors was fixed at one for every instance.
The anomaly scores obtained with PCCA were calculated
based on the reconstruction errors.
HOAD requires to select an appropriate hyperparameter value
for controlling the constraints whereby different views 
of the same instance are embedded close together.
We ran HOAD with different hyperparameter settings $\{0.1,1,10,100\}$,
and show the results that achieved the highest performance
for each data set.
For CC, first we clustered instances for each view 
using spectral clustering~\citep{ng2002spectral}.
We set the number of clusters at 20, 
which achieved a good performance in preliminary experiments.
Then, we calculated anomaly scores 
by the likelihood of consensus clustering
when an instance was removed.
OCSVM is a representative method for single-view anomaly detection.
To investigate the performance of a single-view method
for multi-view anomaly detection, we included OCSVM as a comparison method.
For OCSVM, multiple views are concatenated in a single vector,
then use it for the input. We used Gaussian kernel.
In the proposed model, we used $\gamma=1$, $a=1$, 
and $b=1$ for all experiments.
The number of iterations for the Gibbs sampling was 500,
and the anomaly score was calculated by averaging over the multiple samples.

\subsection{Multi-view anomaly detection}
\begin{figure*}[t!]
\centering
{\tabcolsep=0.2em
\begin{tabular}{ccc}
(a) breast-cancer & (b) diabetes & \\
\includegraphics[height=8em]{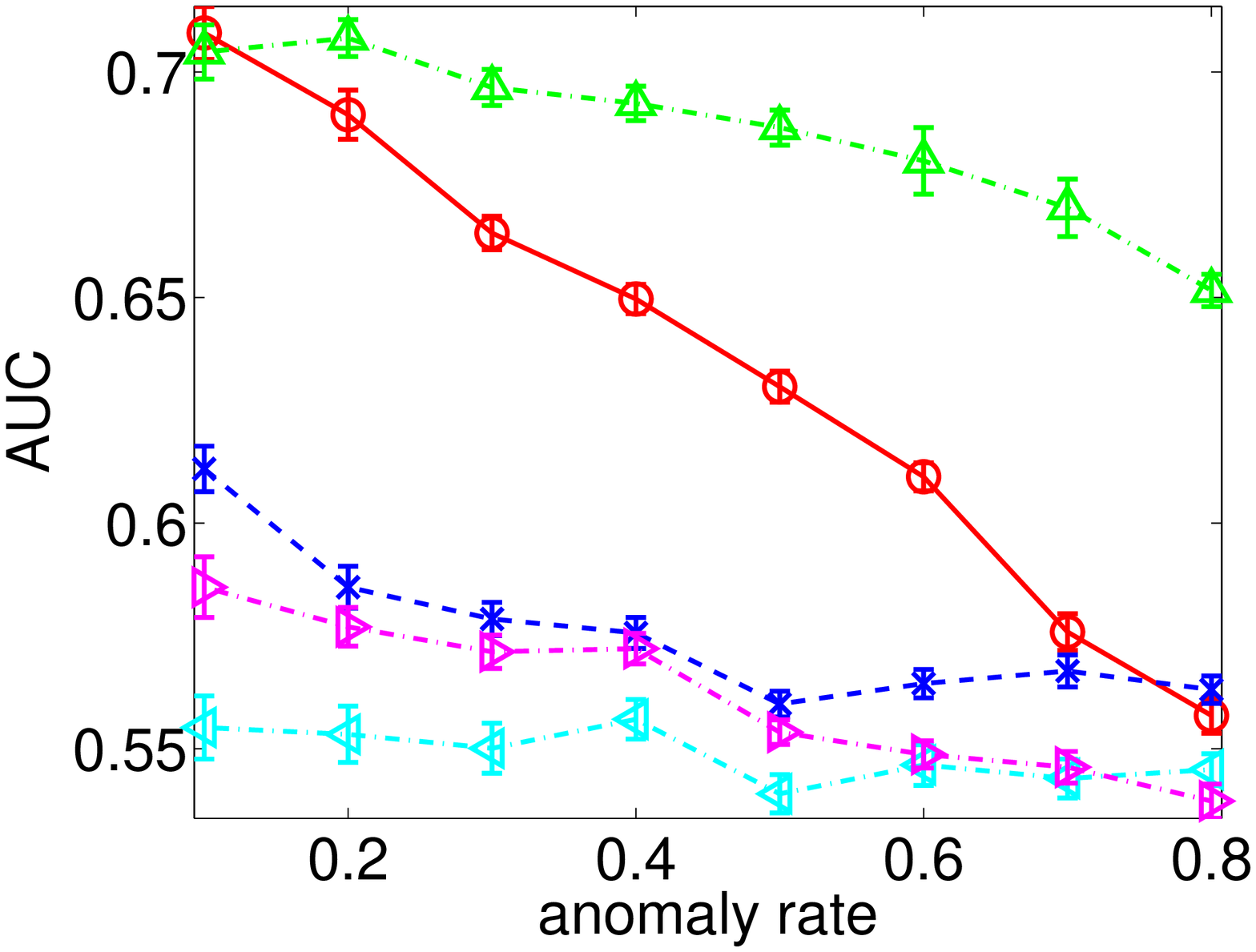} &
\includegraphics[height=8em]{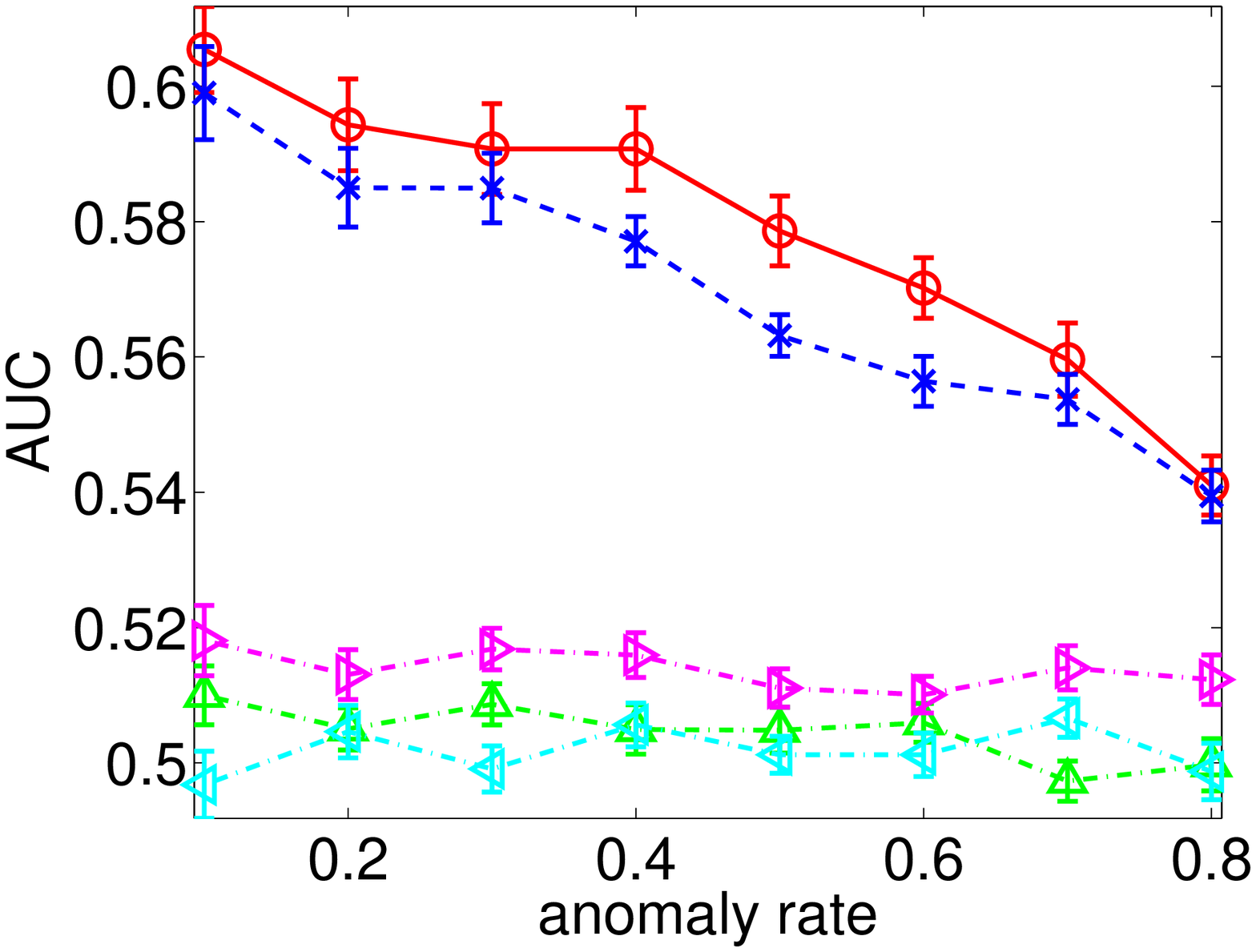} &
\includegraphics[height=8em]{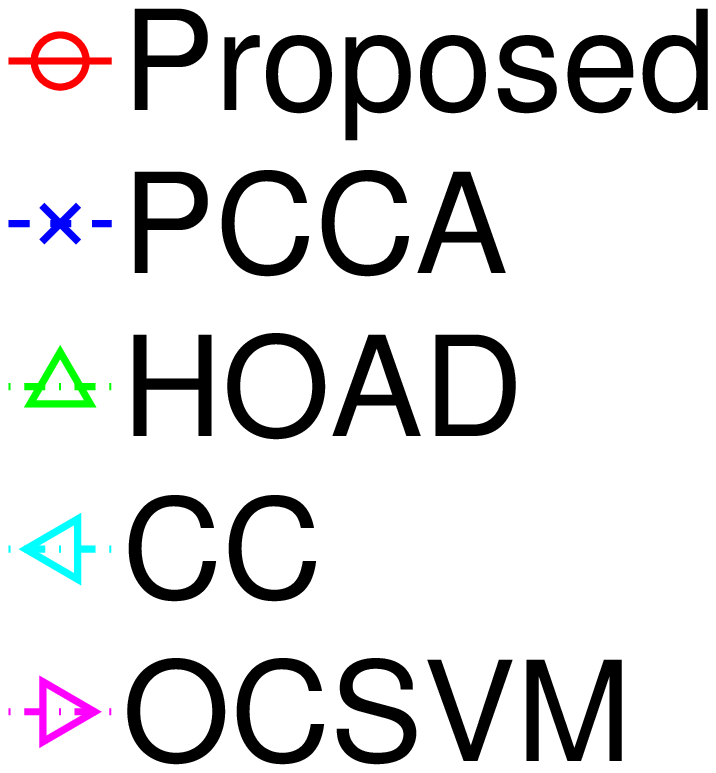}\\
(c) glass & (d) heart & (e) ionosphere \\
\includegraphics[height=8em]{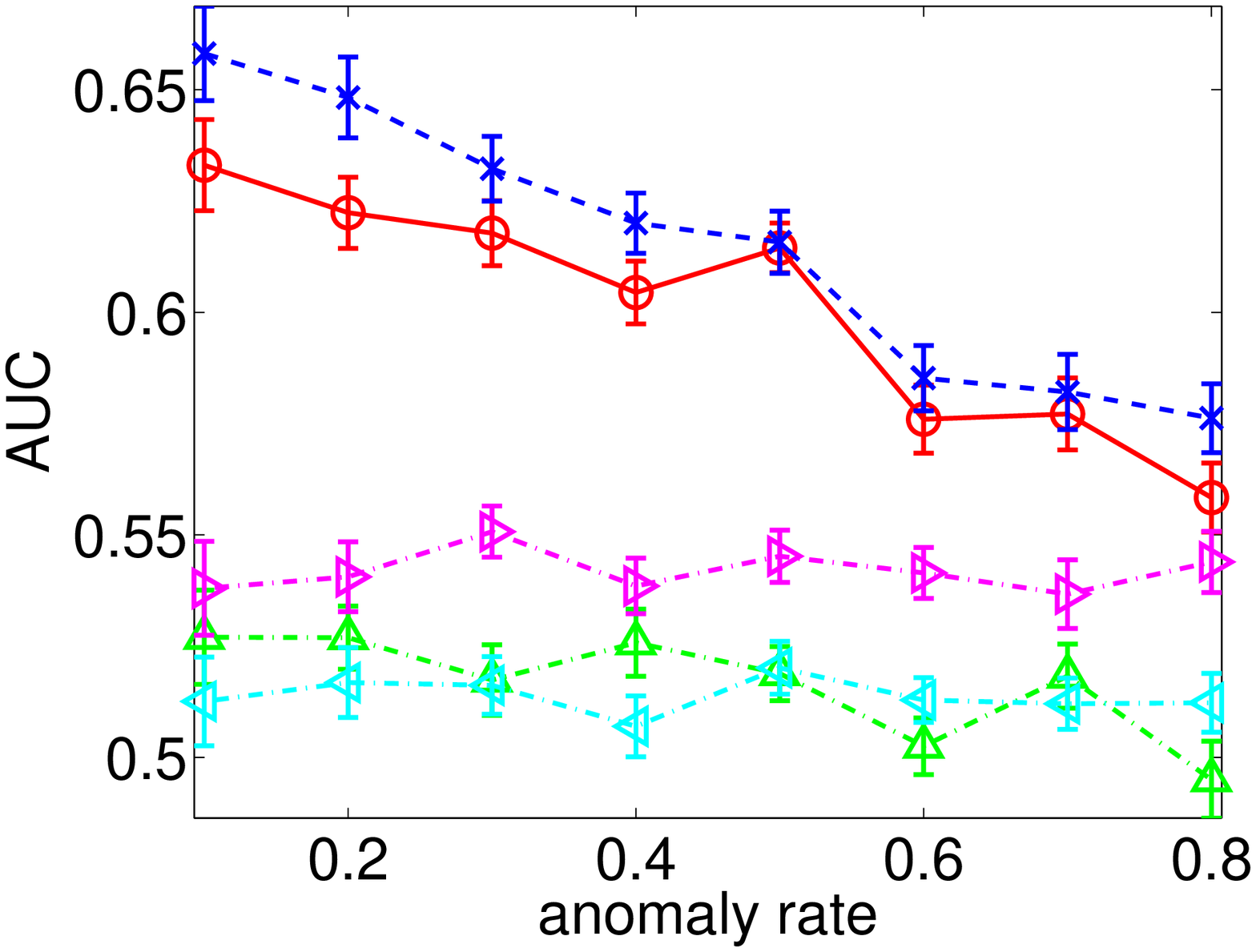} &
\includegraphics[height=8em]{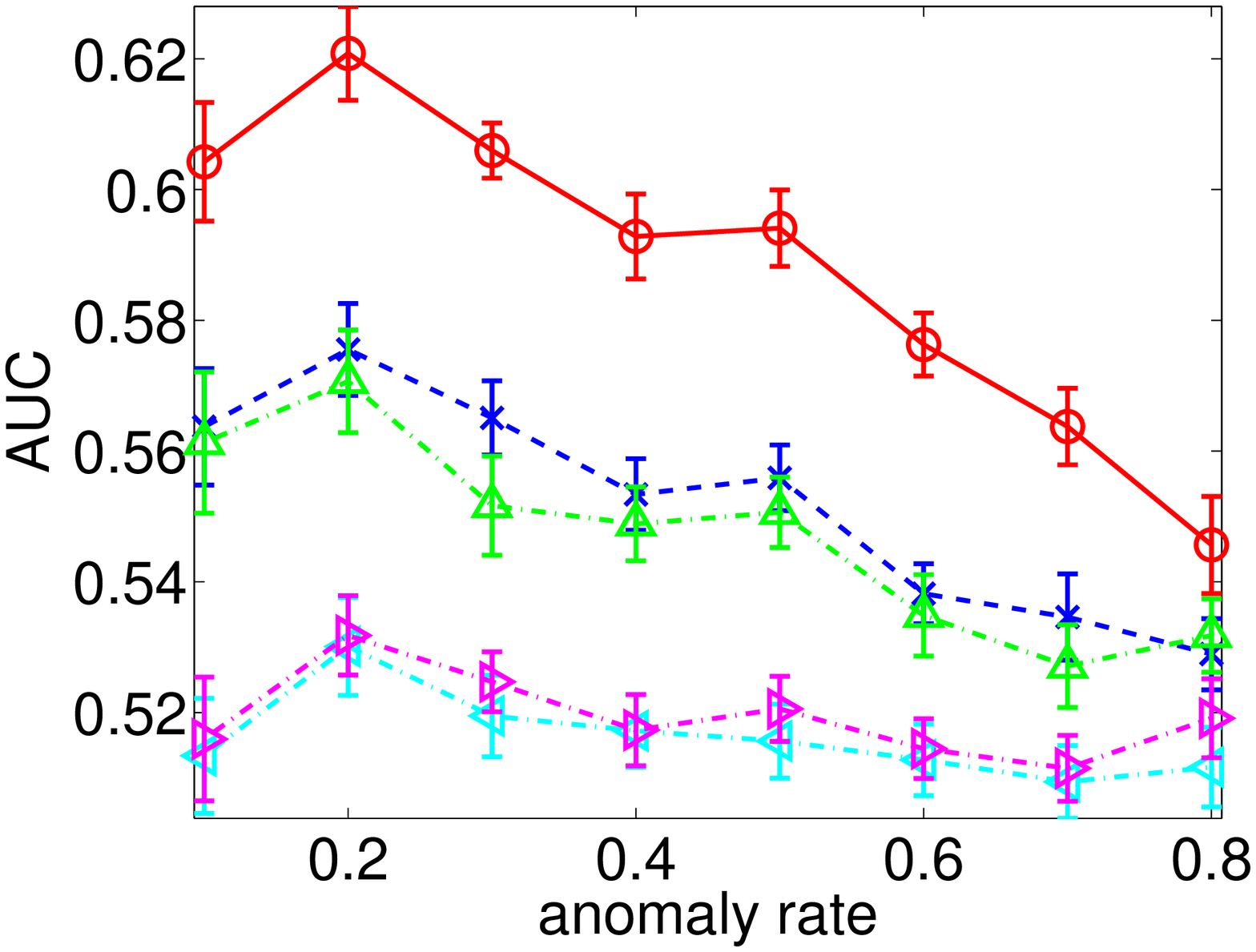} &
\includegraphics[height=8em]{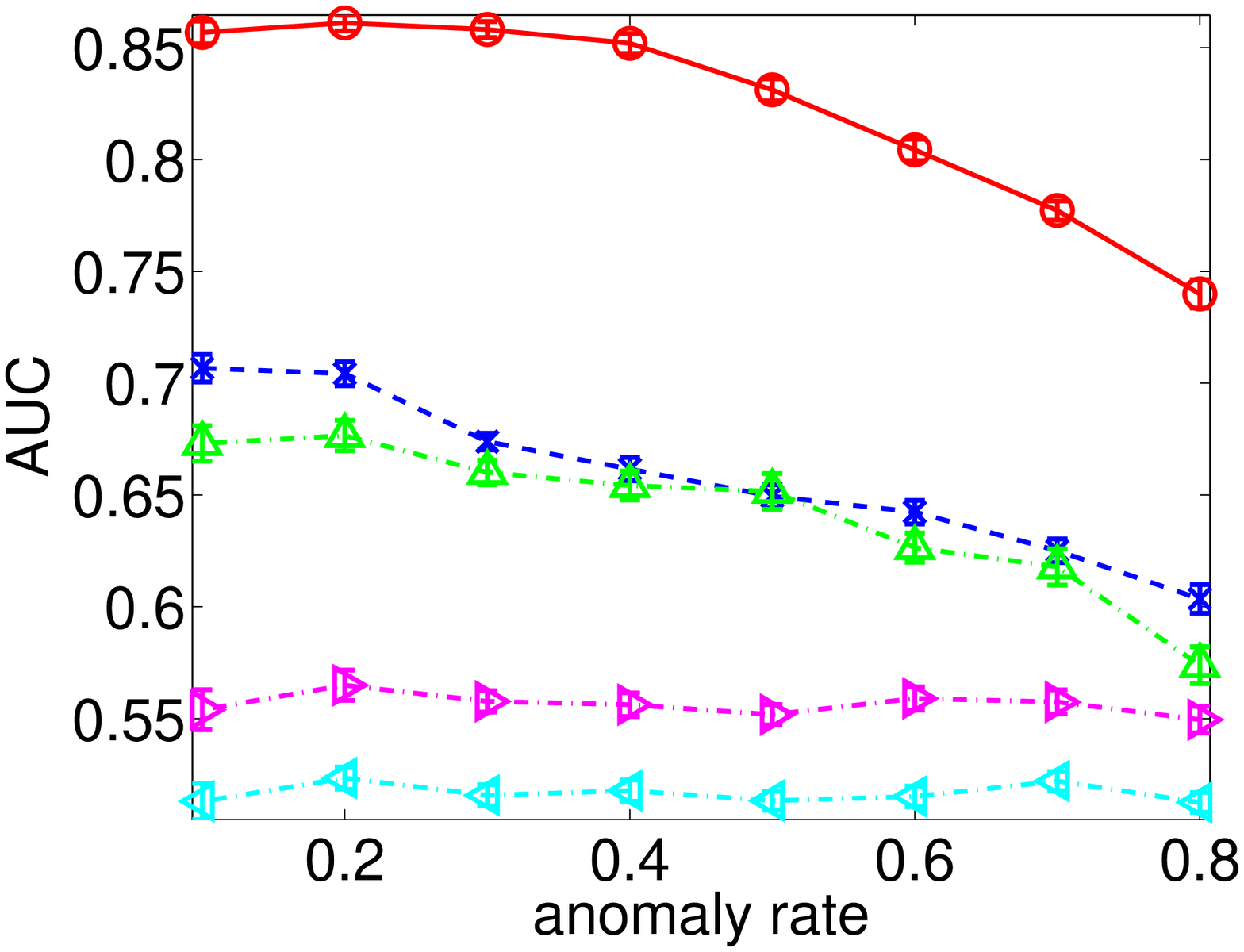} \\
(f) sonar & (g) svmguide2 & (h) svmguide4 \\
\includegraphics[height=8em]{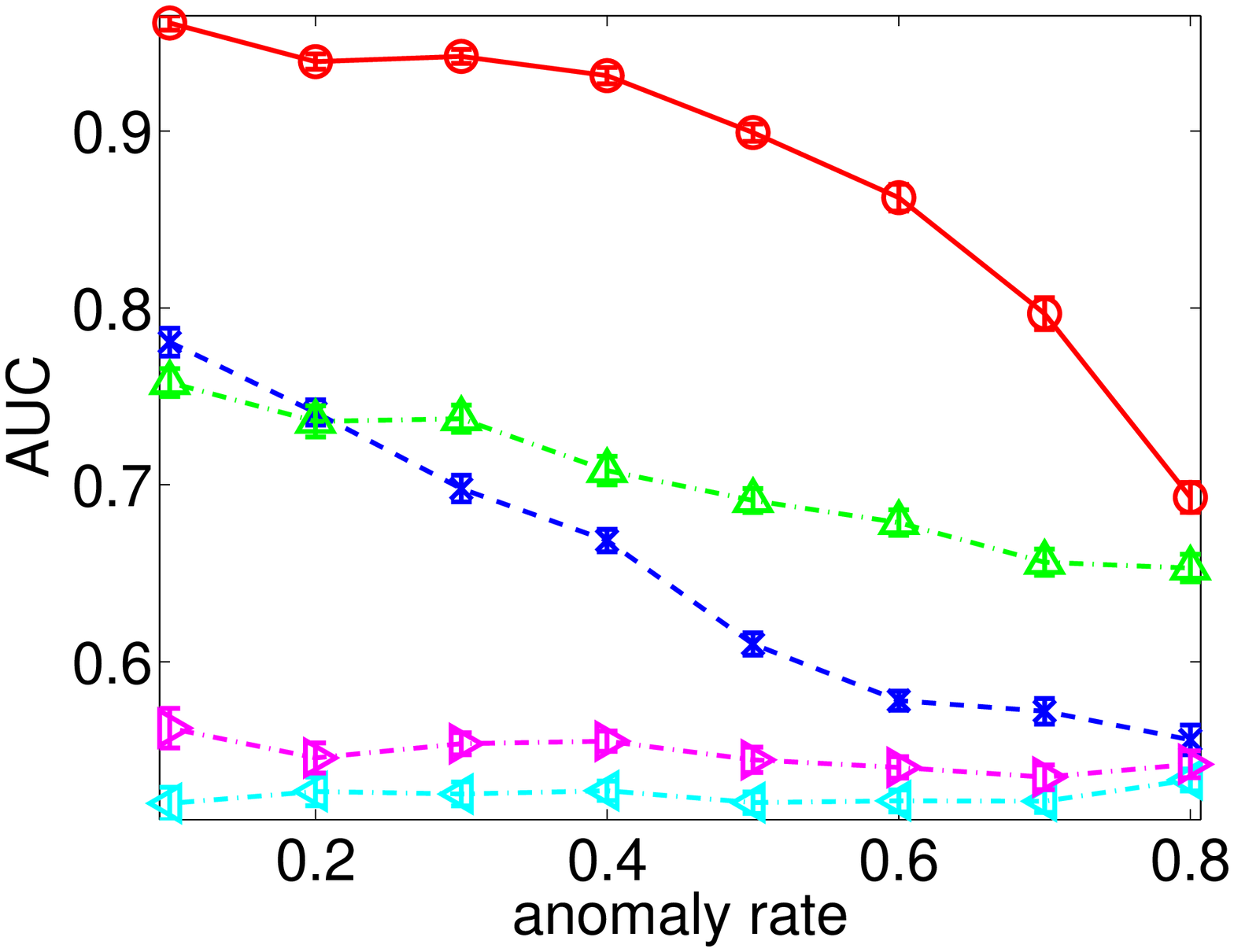} &
\includegraphics[height=8em]{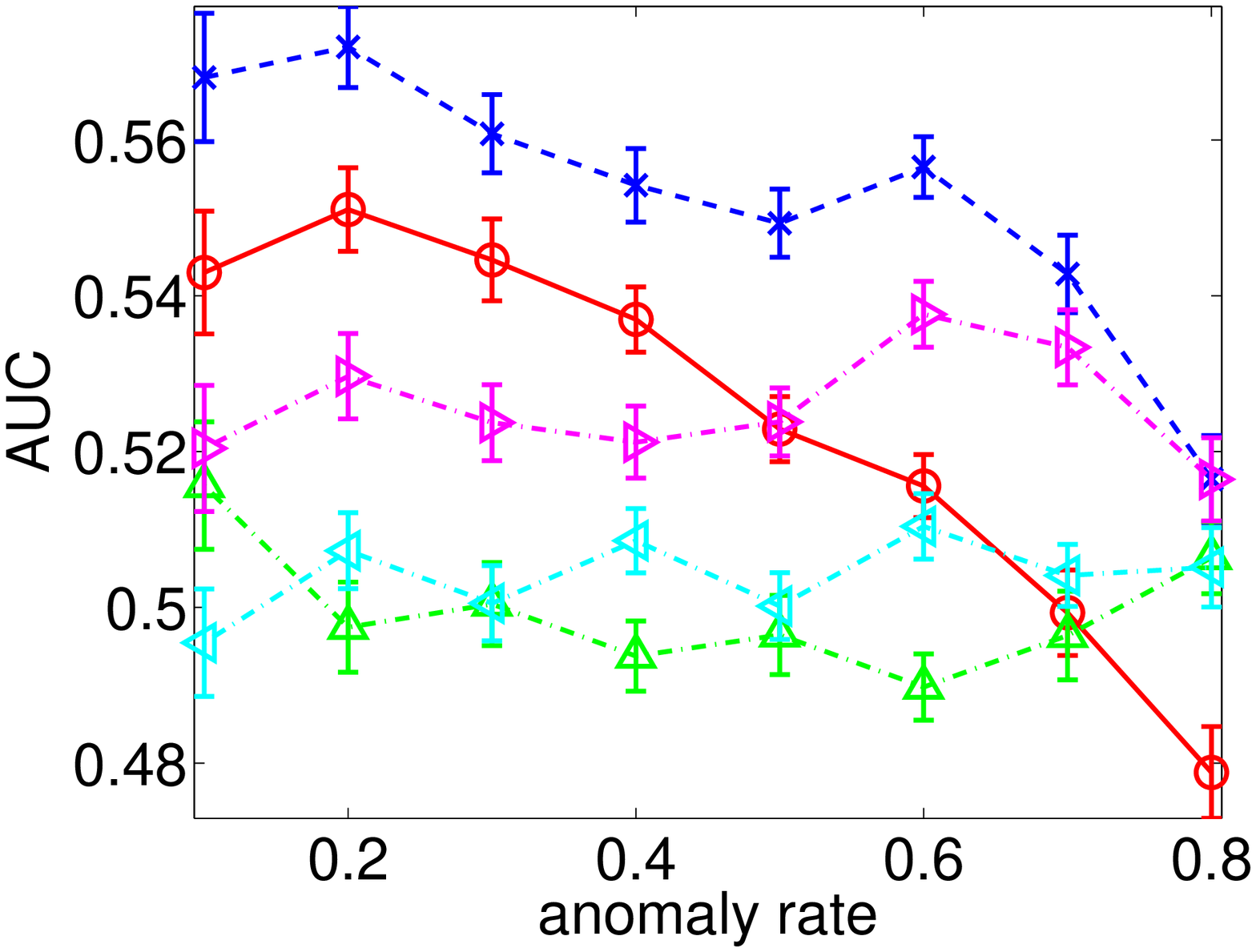} &
\includegraphics[height=8em]{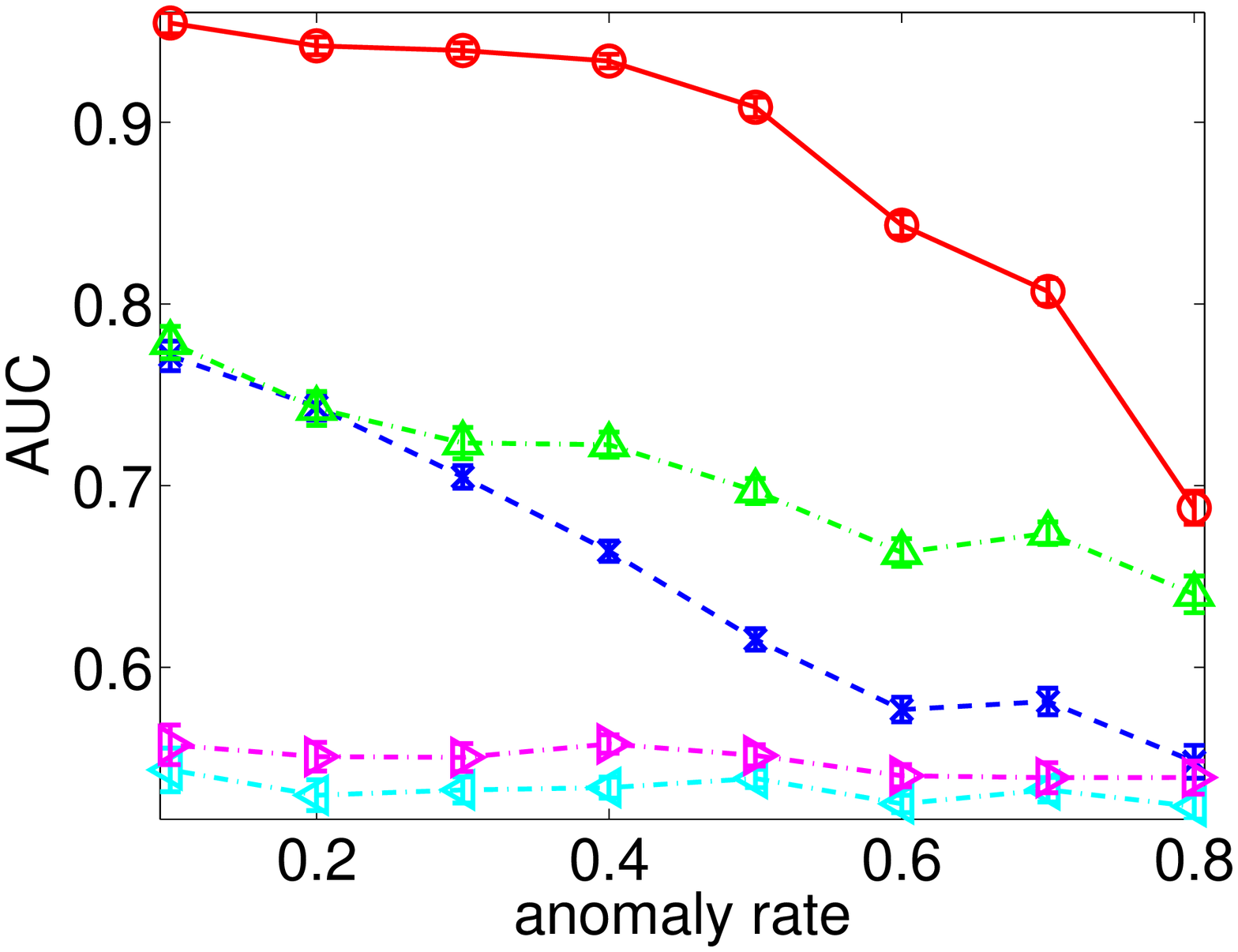} \\
(i) vehicle & (j) vowel & (k) wine \\
\includegraphics[height=8em]{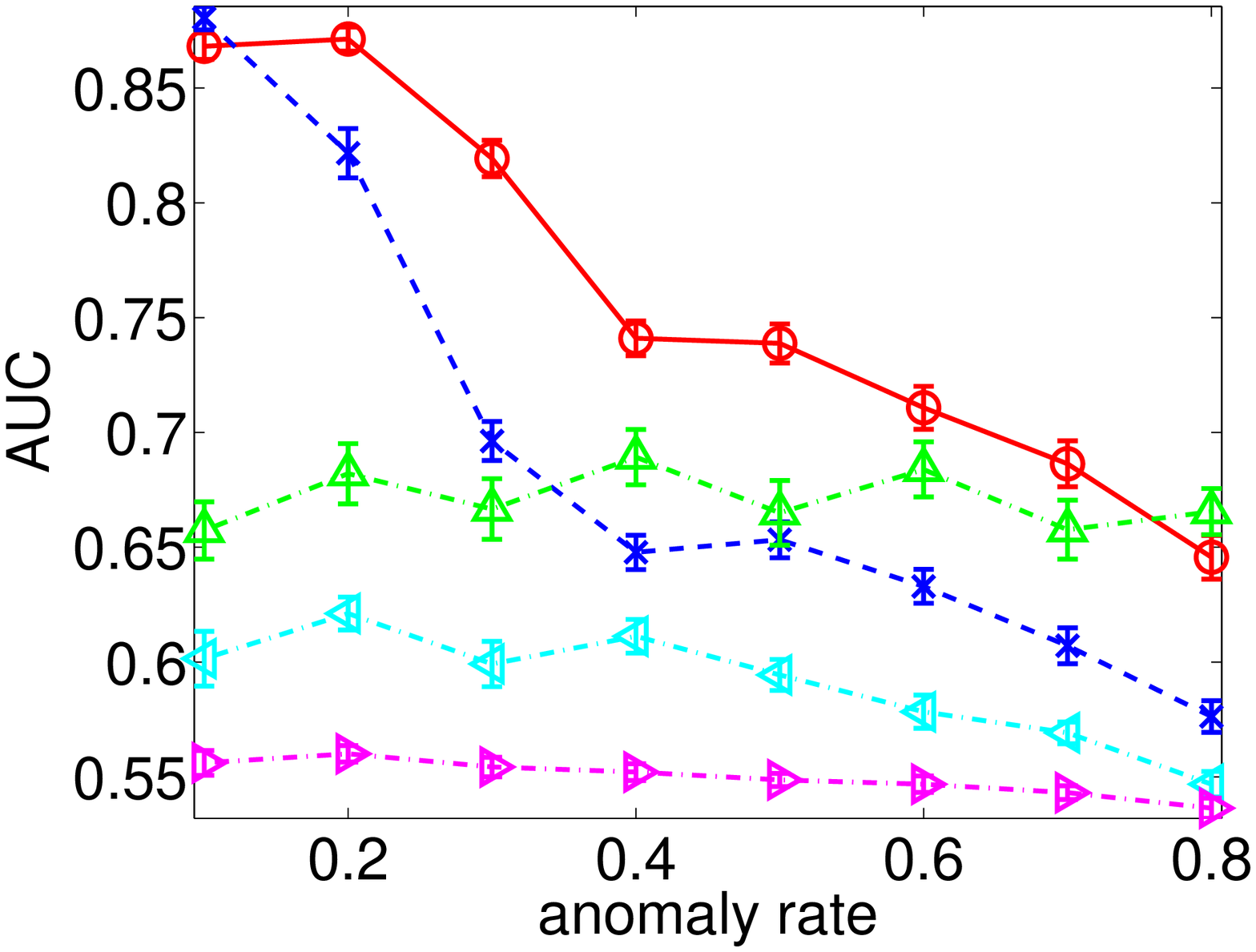} &
\includegraphics[height=8em]{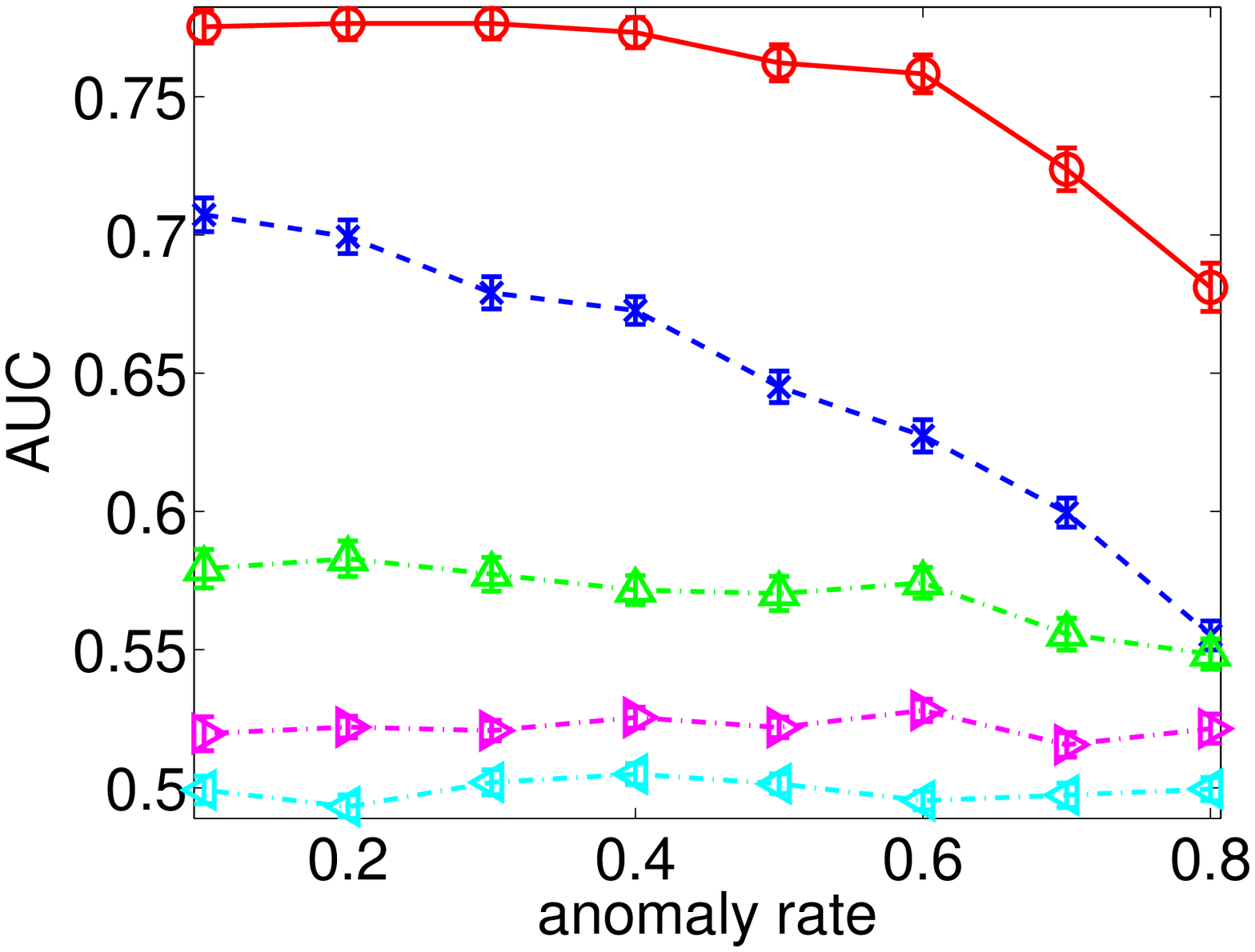} &
\includegraphics[height=8em]{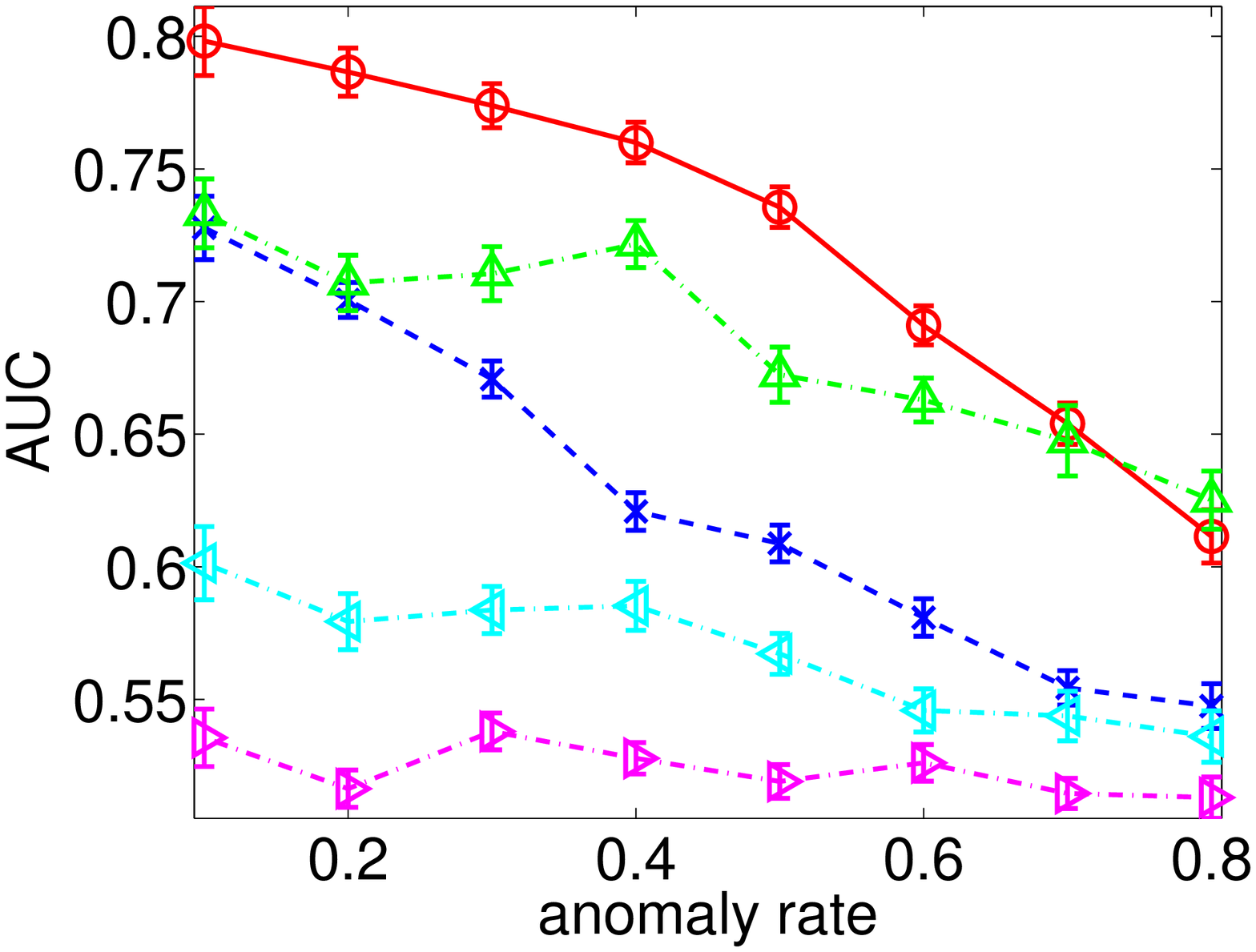}
\\
\end{tabular}}
\caption{Average AUCs with different anomaly rates, and their standard errors. A higher AUC is better.}
\label{fig:auc_rate}
\end{figure*}

For the evaluation measurement, we used the area under the ROC curve (AUC).
A higher AUC indicates a higher anomaly detection performance.
Figure~\ref{fig:auc_rate} shows AUCs with different rates of anomalies
using two-view data sets,
which are averaged over 50 experiments. 
For the dimensionality of the latent space, we used $K=5$ for 
the proposed model, PCCA, and HOAD.
In general, as the anomaly rate increases, the performance decreases.
The proposed model achieved the best performance with 
eight of the 11 data sets. 
This result indicates that the proposed model
can find anomalies effectively 
by inferring a number of latent vectors for each instance.
The performance of CC was low because it assumes that
there are clusters for each view, and it cannot find 
anomalies within clusters.
The AUC of OCSVM was low, because it is 
a single-view anomaly detection method,
which considers instances anomalous
that are different from others within a single view.
Multi-view anomaly detection is the task to find
instances that have inconsistent features across views,
but not inconsistent features within a view.
The computational time needed for PCCA was 2 sec, 
and that needed for the proposed model was 35 sec with wine data.

\begin{figure*}[t!]
\centering
{\tabcolsep=0.2em
\begin{tabular}{ccc}
(a) breast-cancer & (b) diabetes & \\ 
\includegraphics[height=8em]{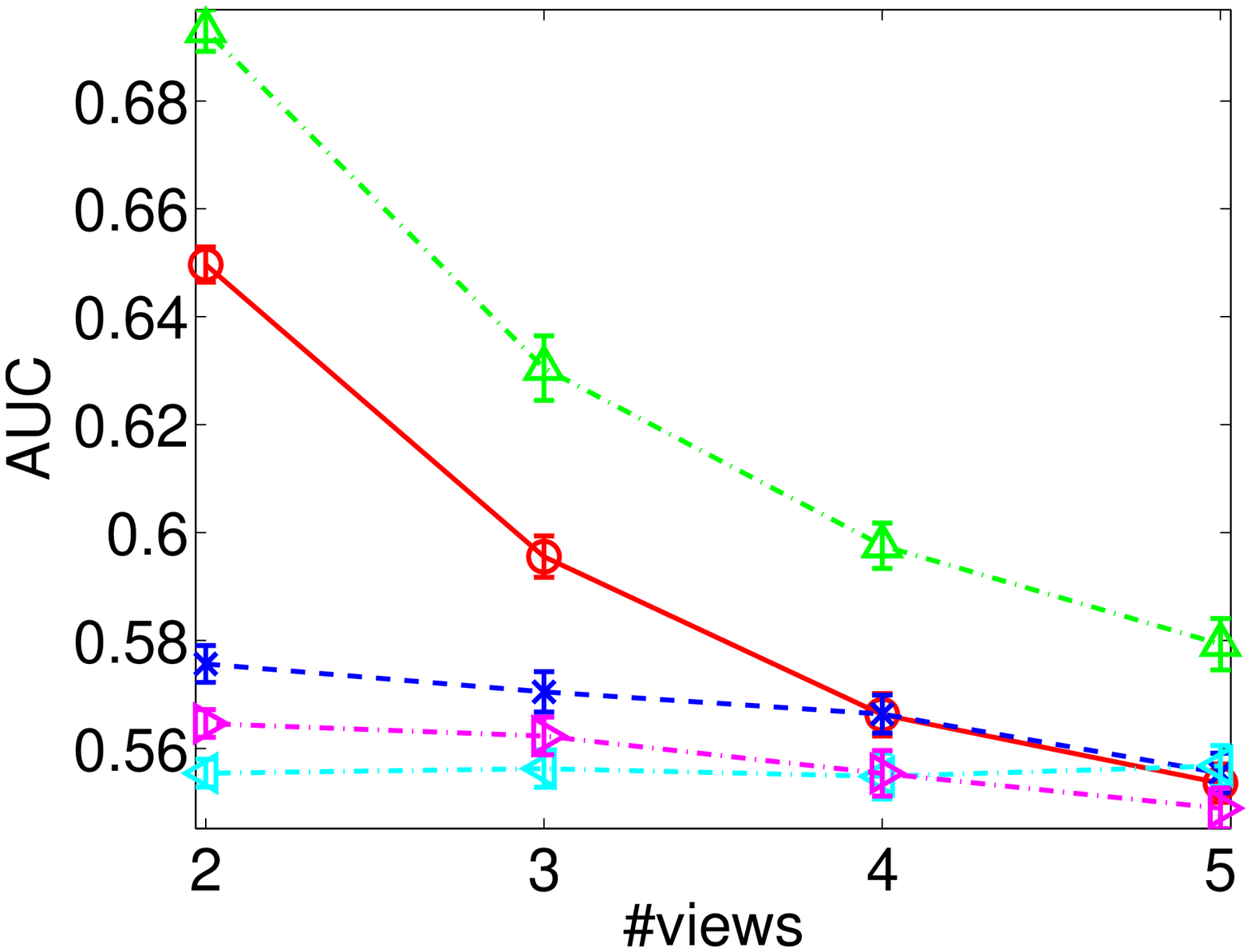} &
\includegraphics[height=8em]{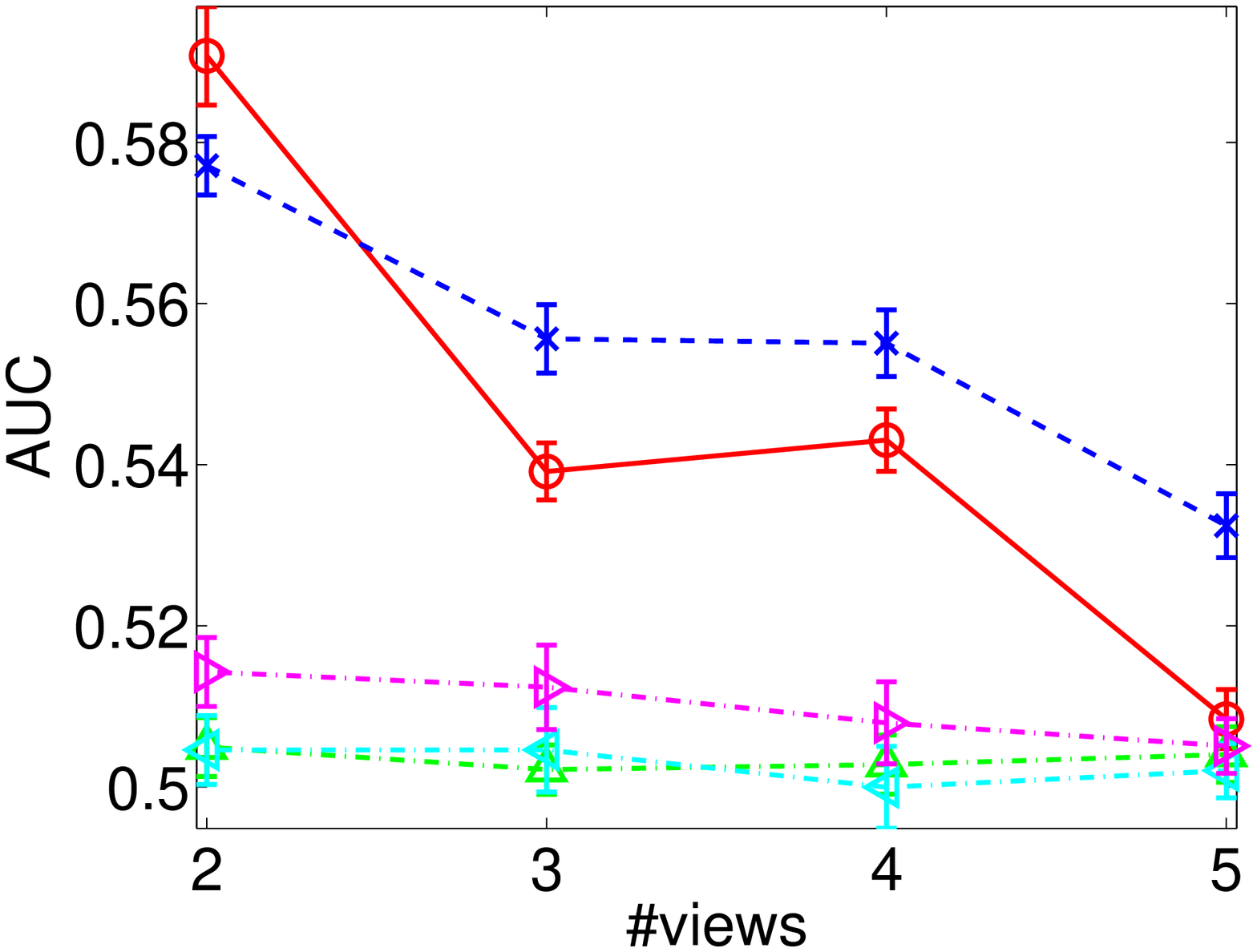} &
\includegraphics[height=8em]{images/icml_auc_legend.eps} \\
(c) glass & (d) heart & (e) ionosphere \\
\includegraphics[height=8em]{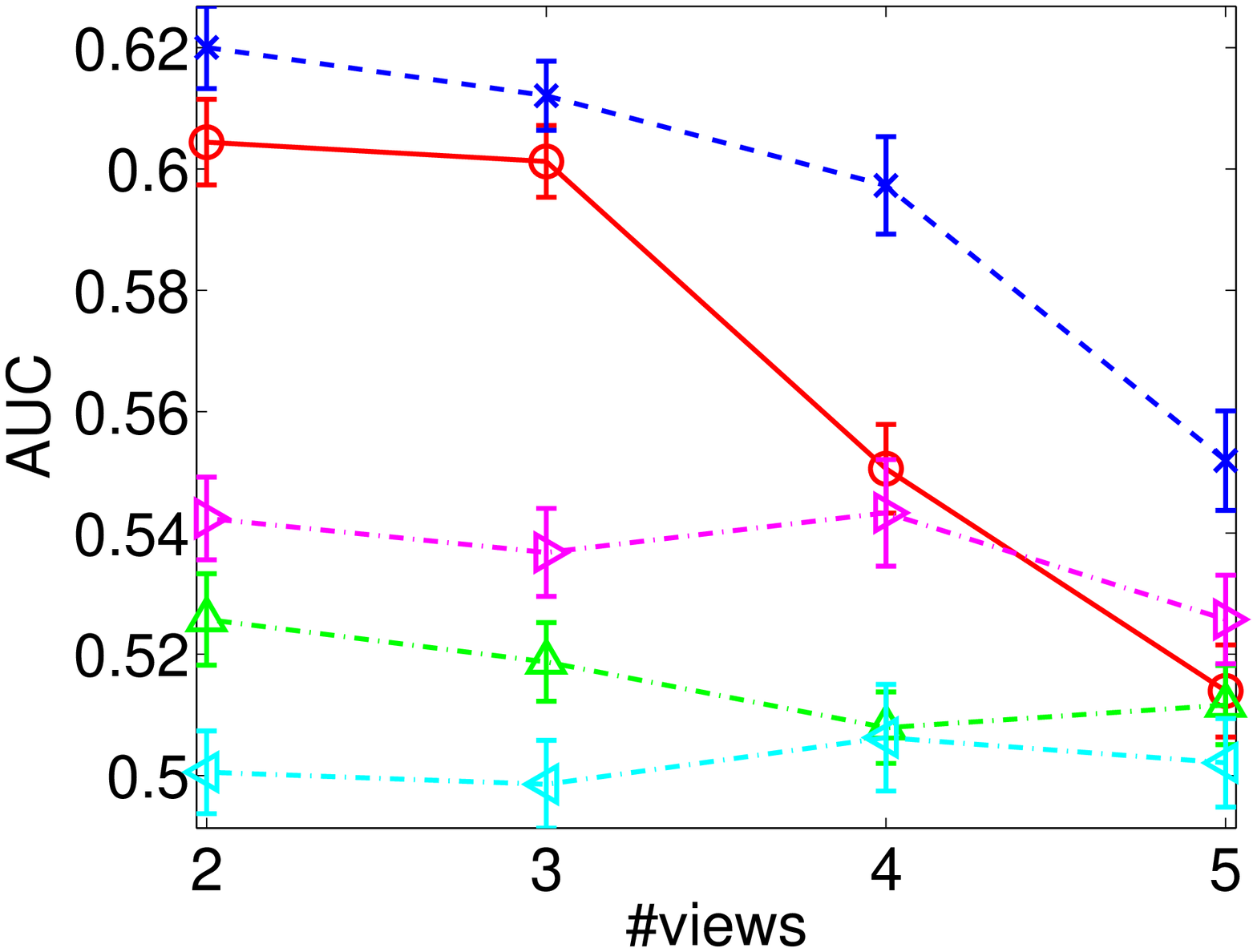} &
\includegraphics[height=8em]{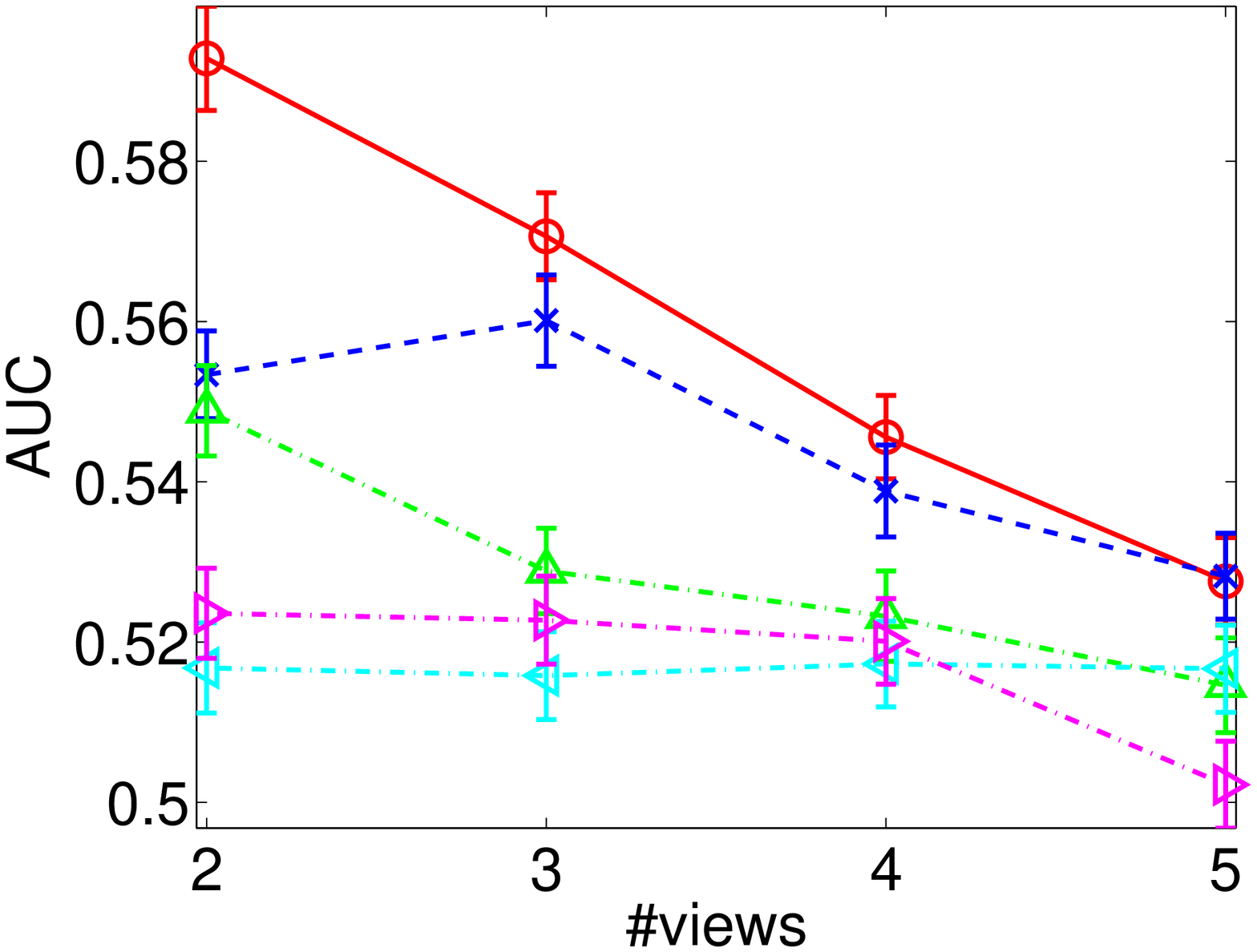} &
\includegraphics[height=8em]{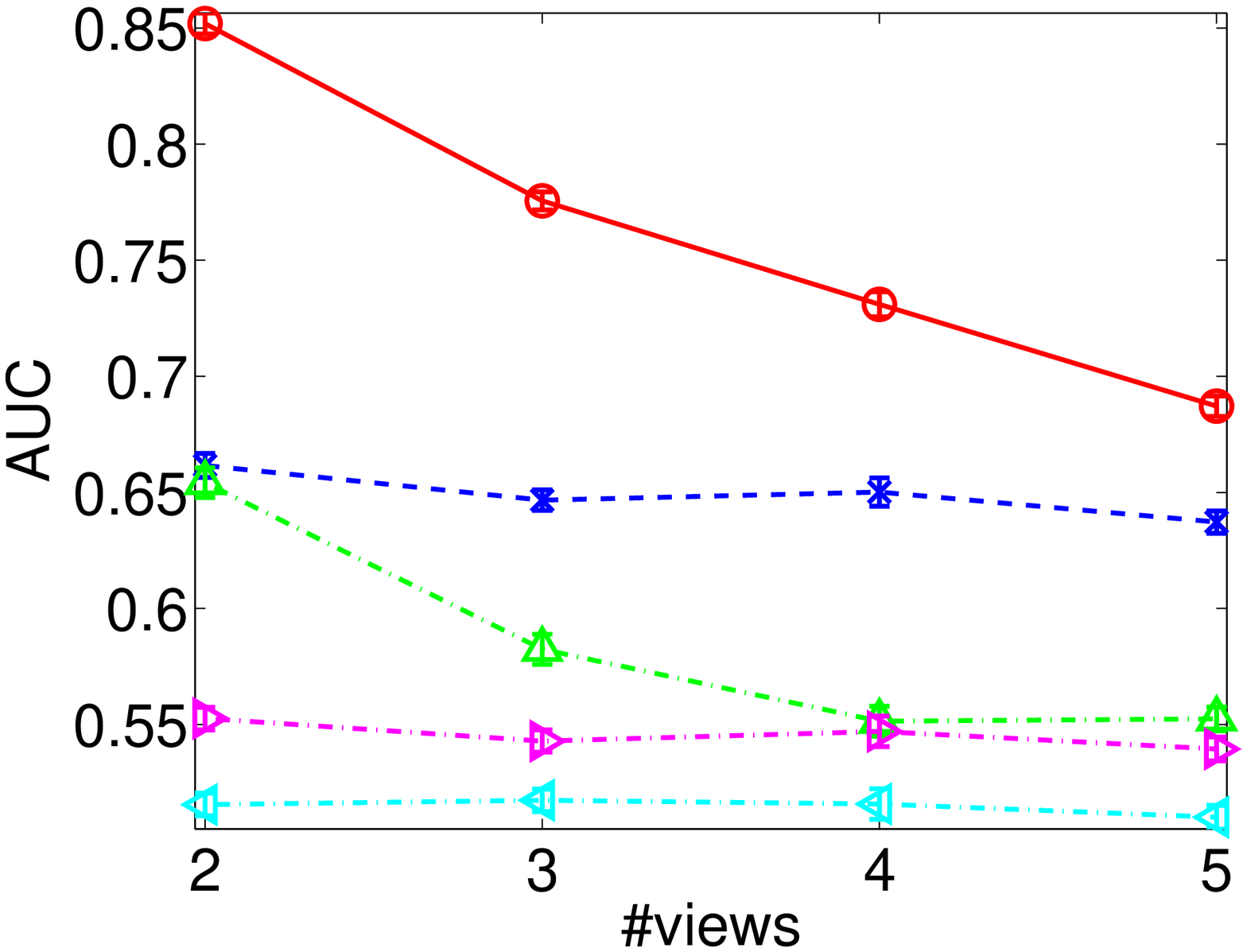} \\
(f) sonar & (g) svmguide2 & (h) svmguide4 \\ 
\includegraphics[height=8em]{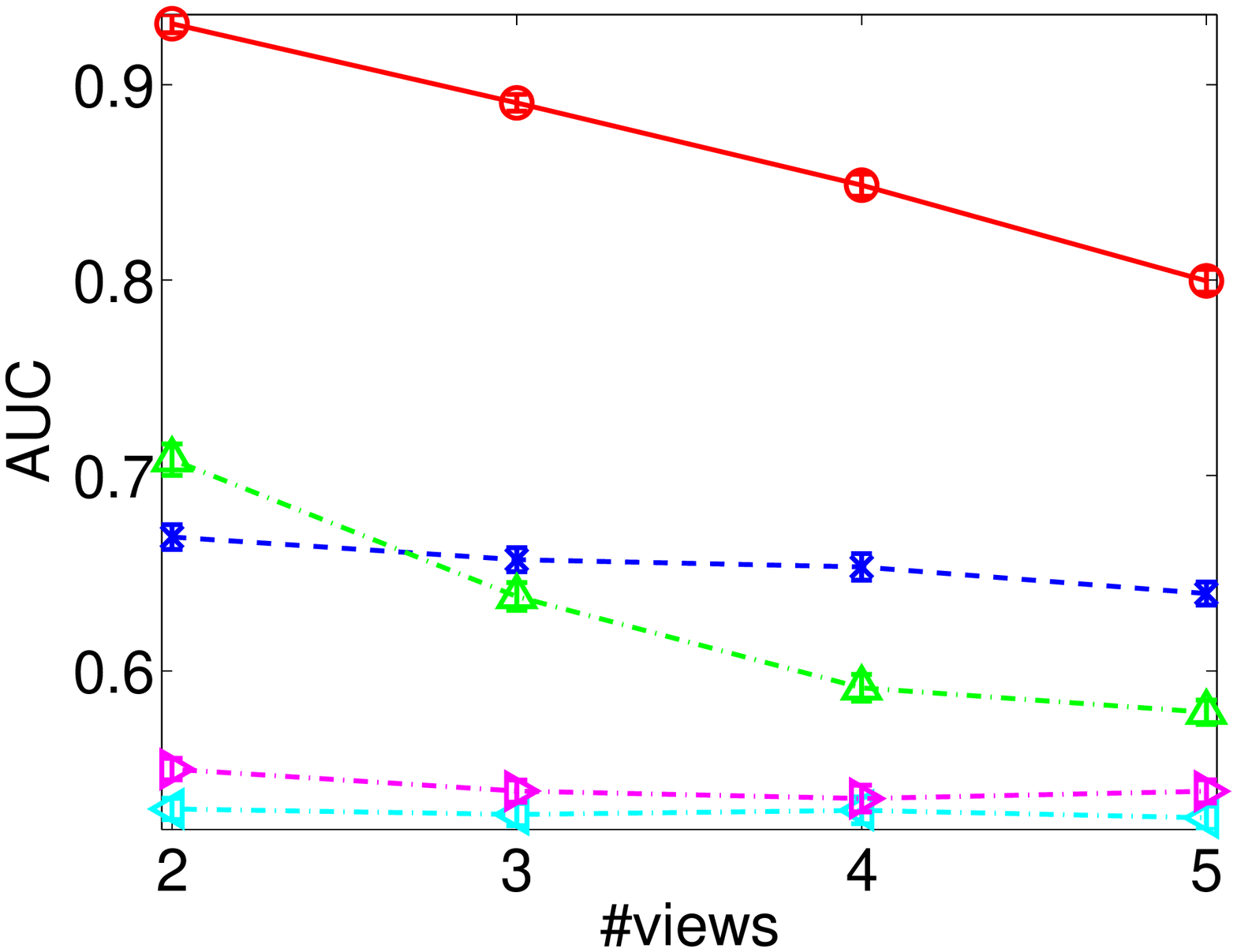} &
\includegraphics[height=8em]{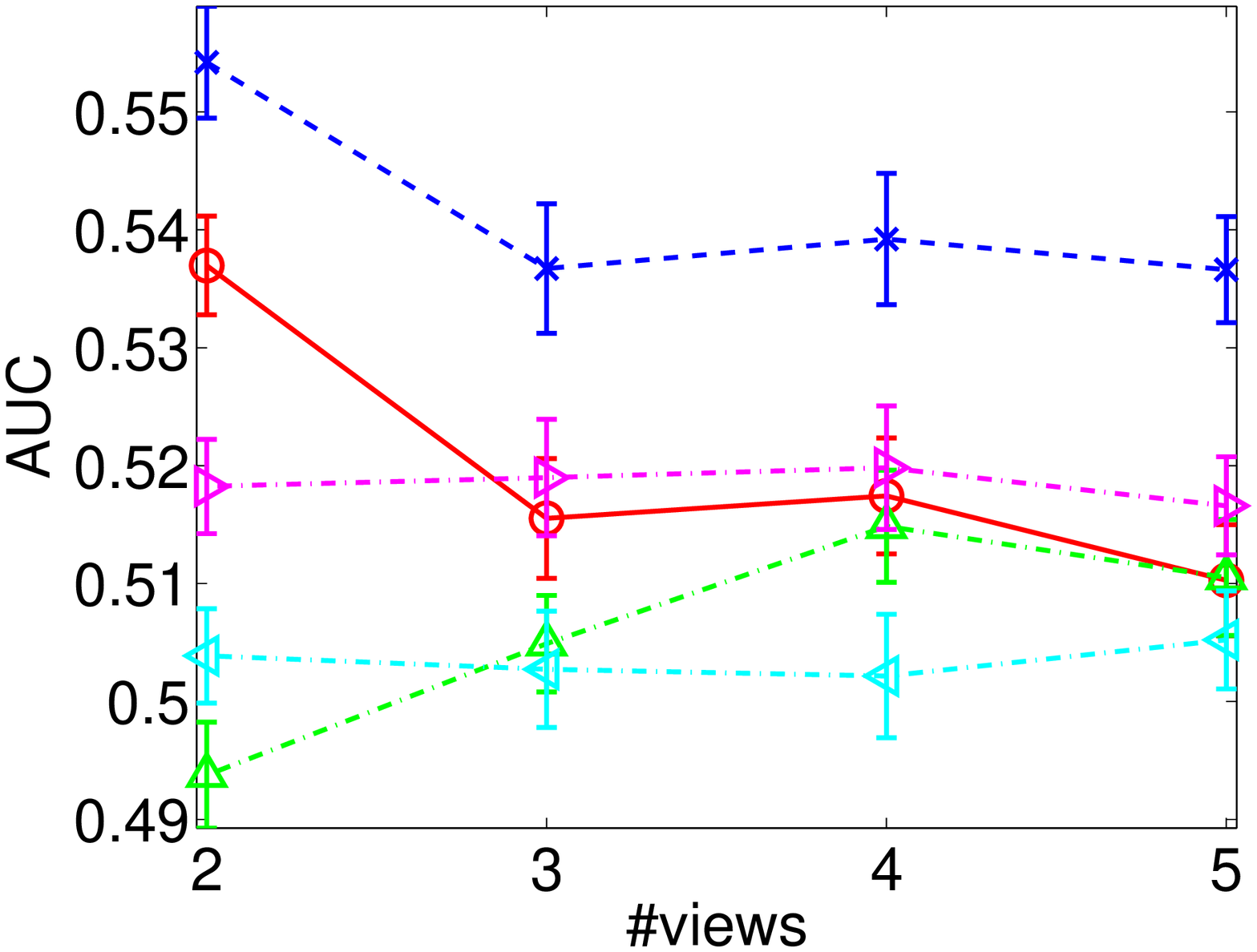} &
\includegraphics[height=8em]{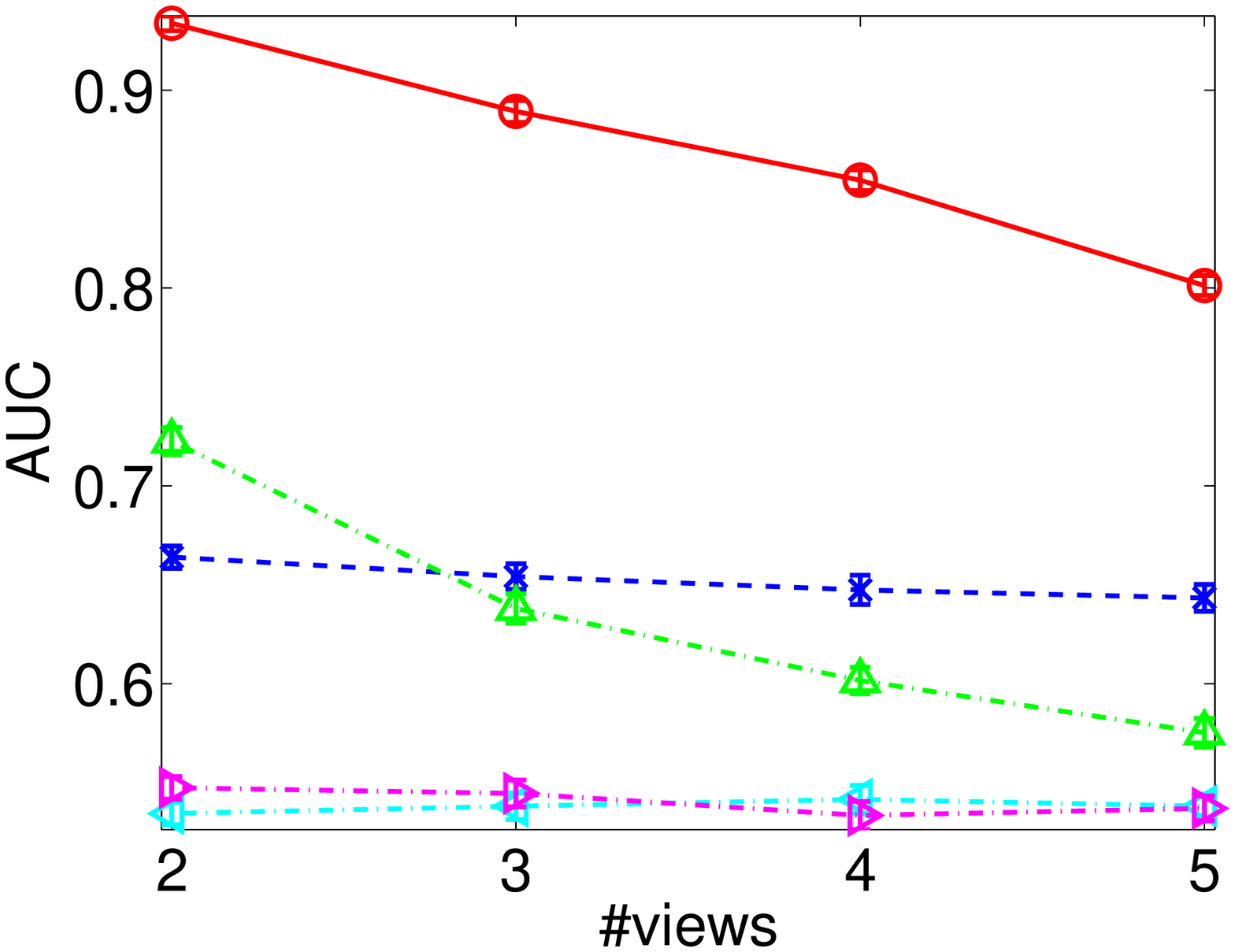} \\
(i) vehicle & (j) vowel & (k) wine \\
\includegraphics[height=8em]{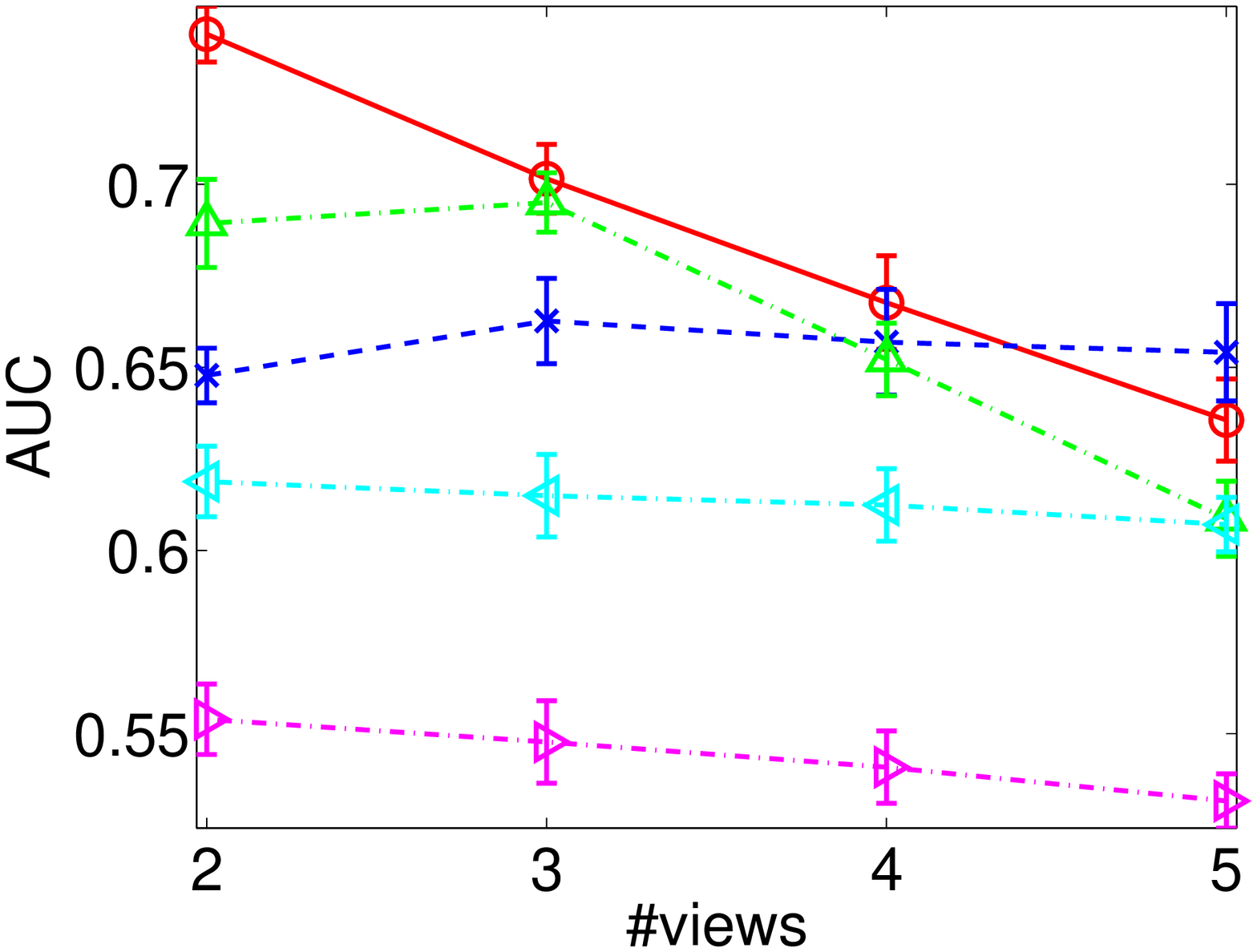} &
\includegraphics[height=8em]{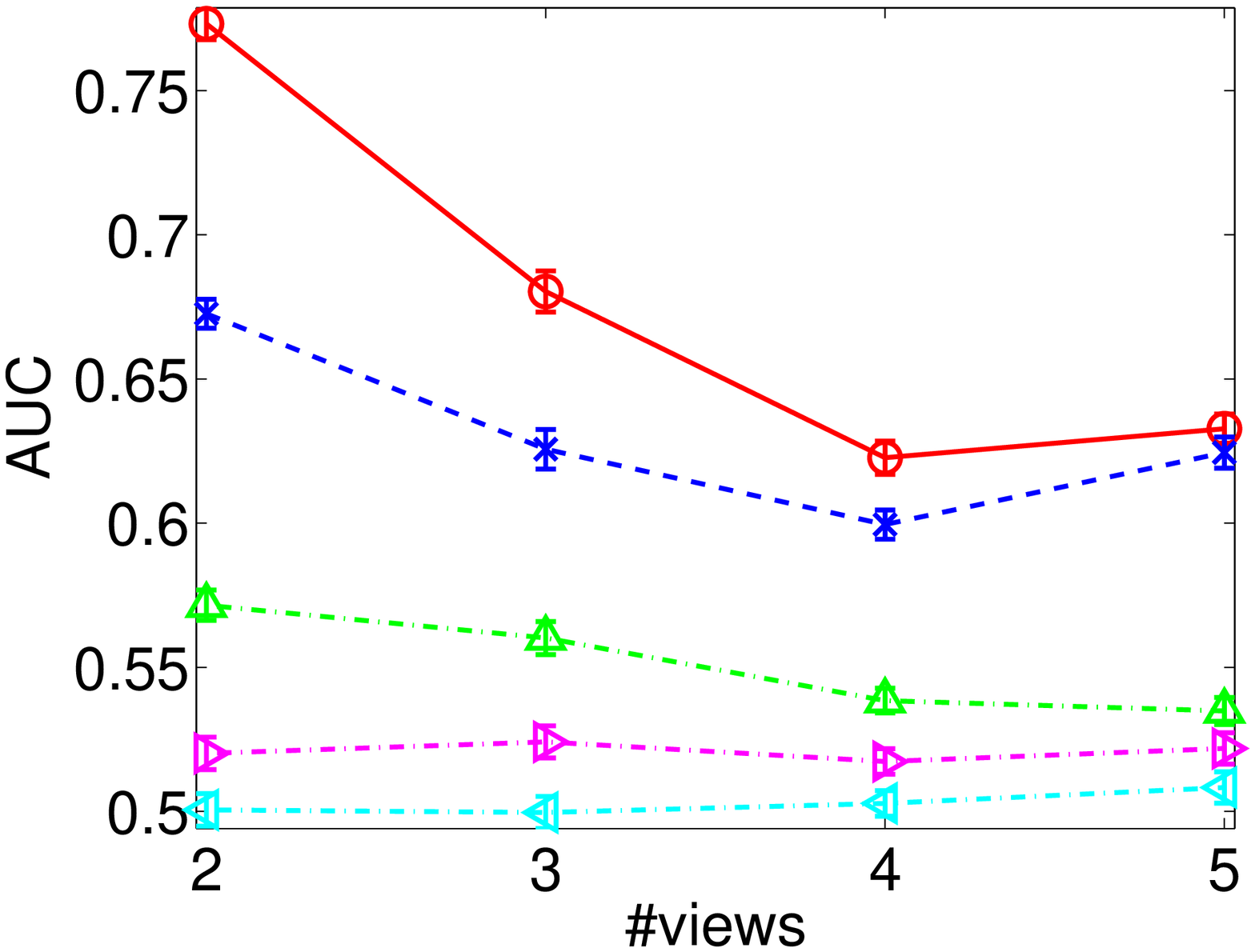} &
\includegraphics[height=8em]{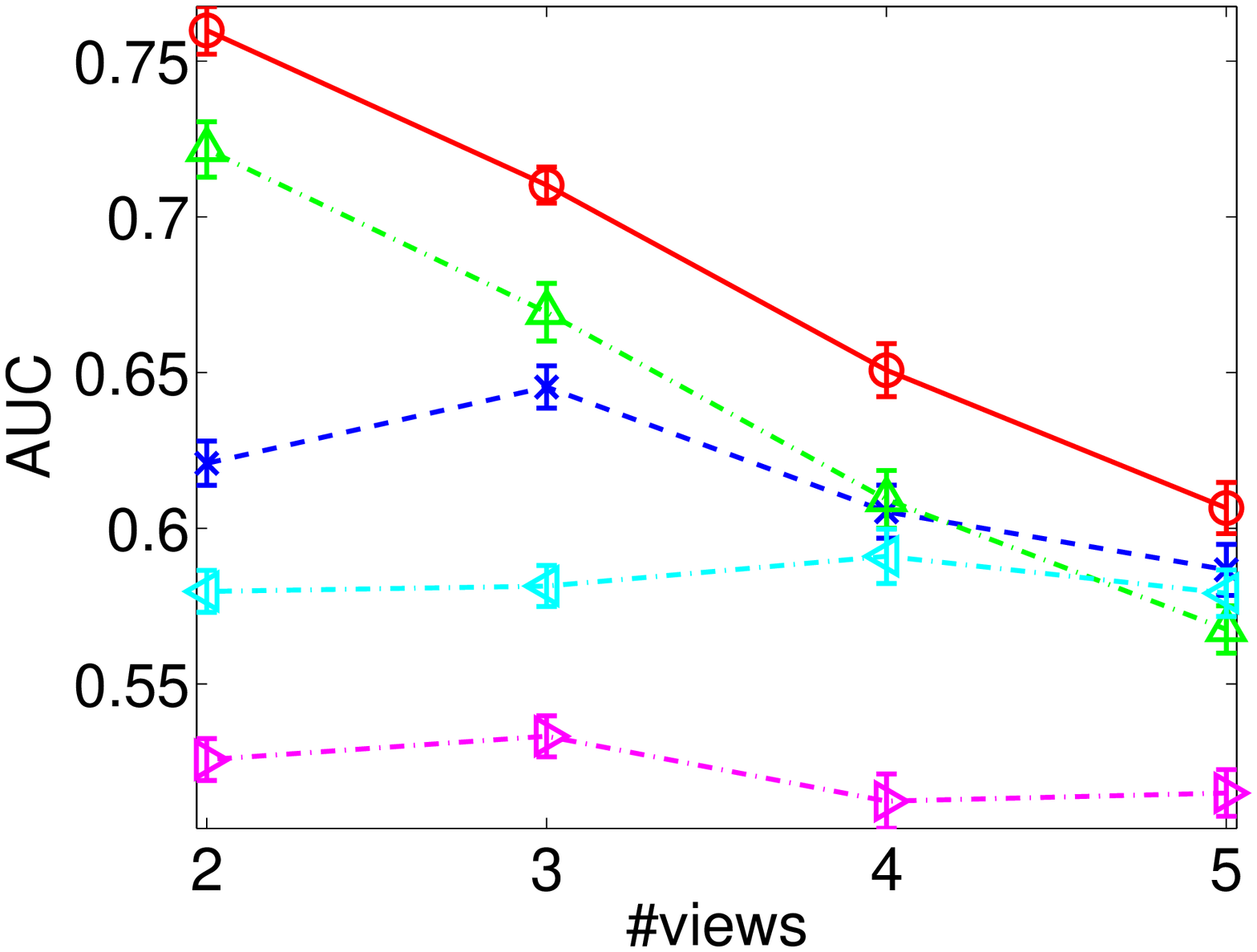} \\
\end{tabular}}
\caption{Average AUCs with different numbers of views, and their standard errors.}
\label{fig:auc_D}
\end{figure*}

Figure~\ref{fig:auc_D} shows AUCs with different numbers of views.
For many cases with different latent dimensionalities 
and different numbers of views,
the proposed model achieved the best performance.

\begin{figure*}[t!]
\centering
{\tabcolsep=0.2em
\begin{tabular}{ccc}
(a) breast-cancer & (b) diabetes & \\
\includegraphics[height=8em]{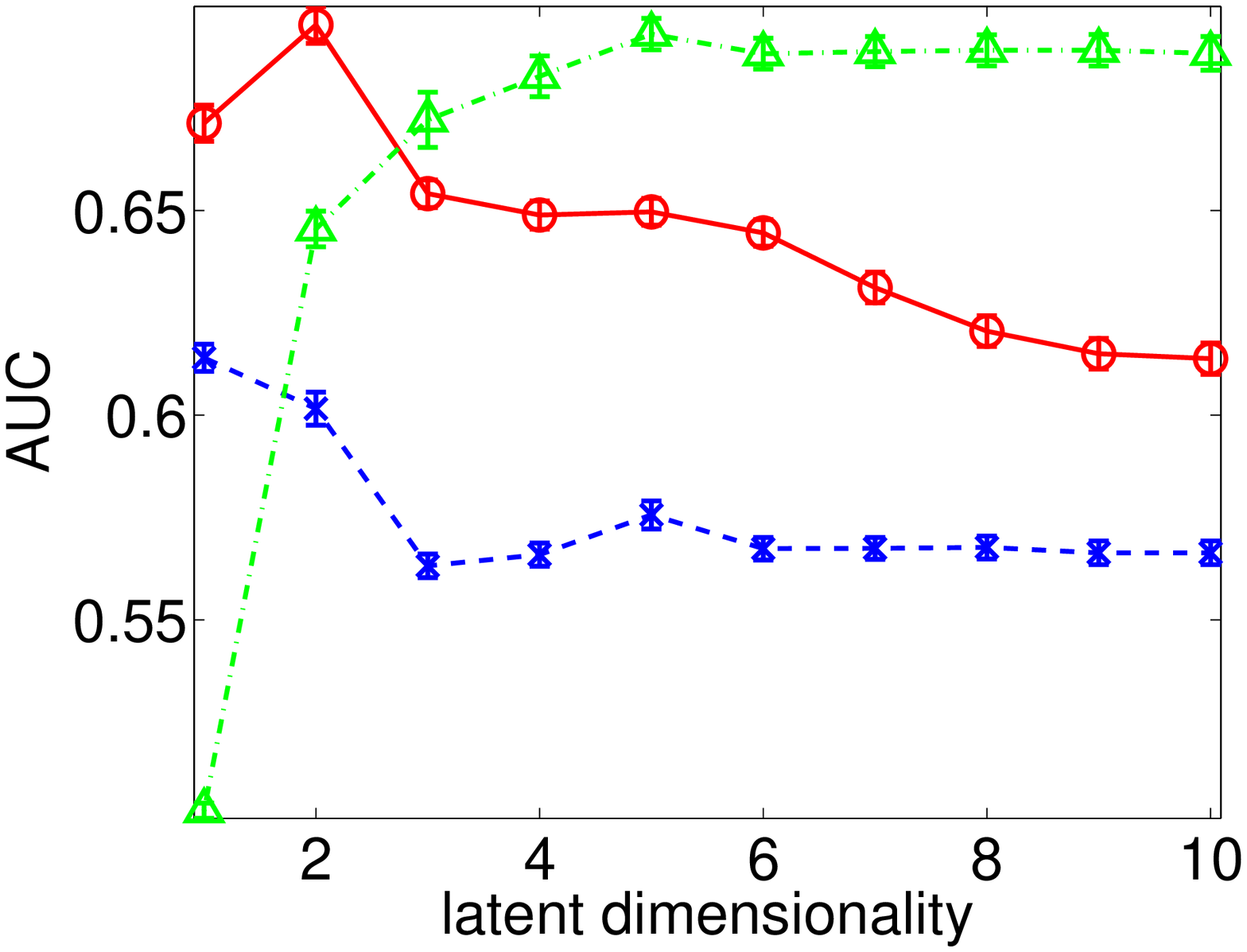} &
\includegraphics[height=8em]{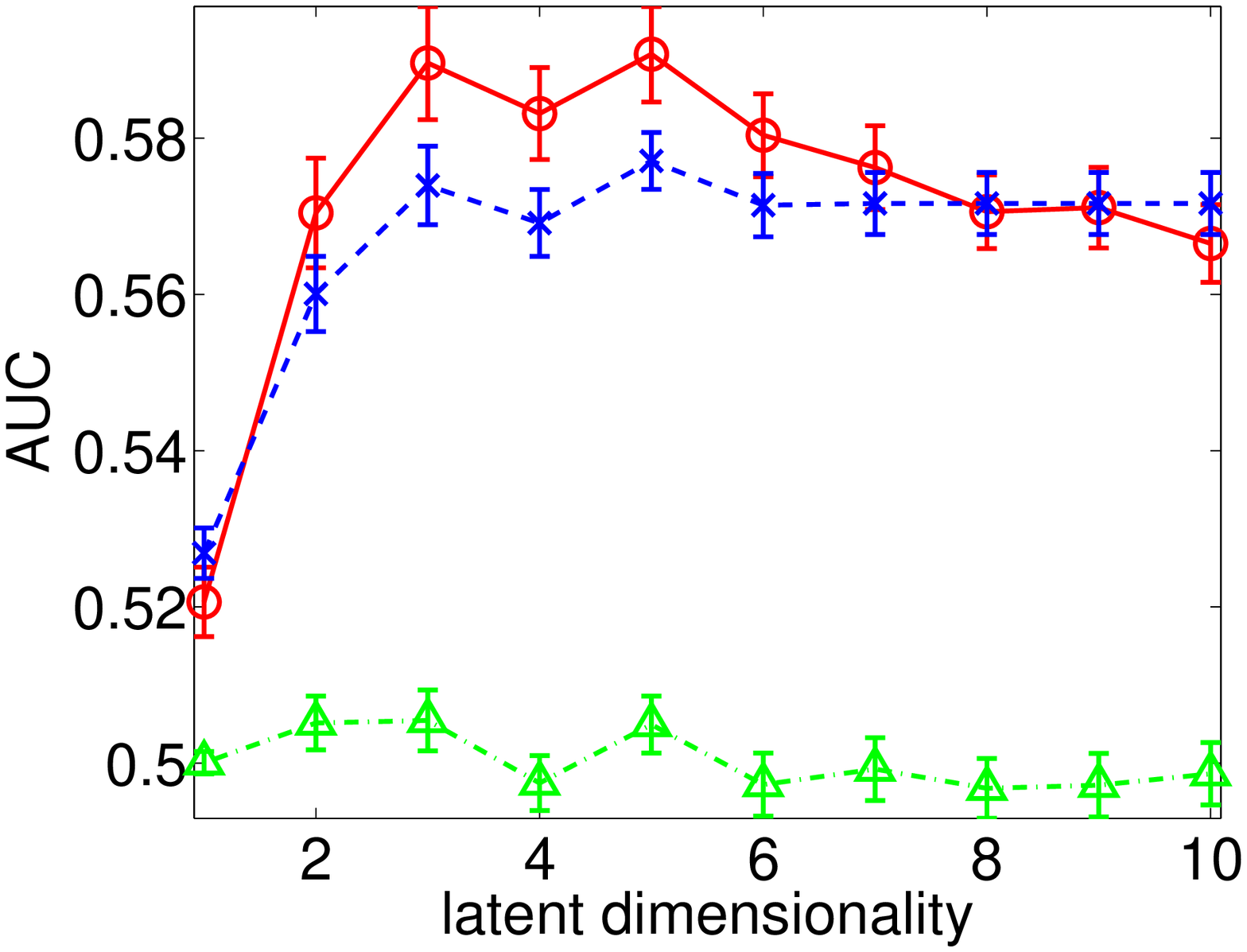} &
\includegraphics[height=8em]{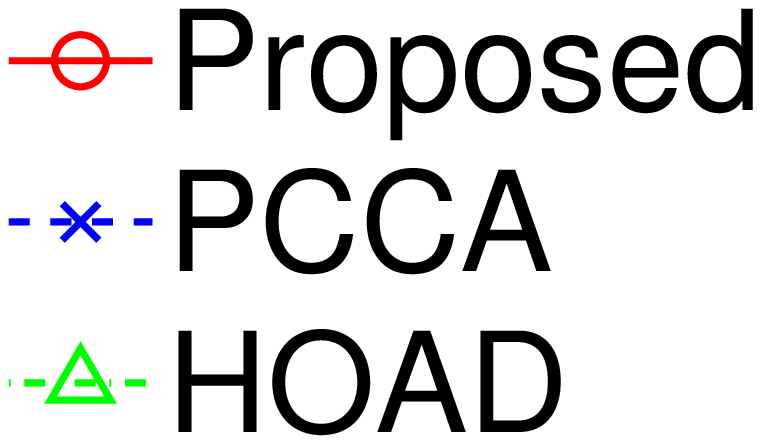} \\
(c) glass & (d) heart & (e) ionosphere \\
\includegraphics[height=8em]{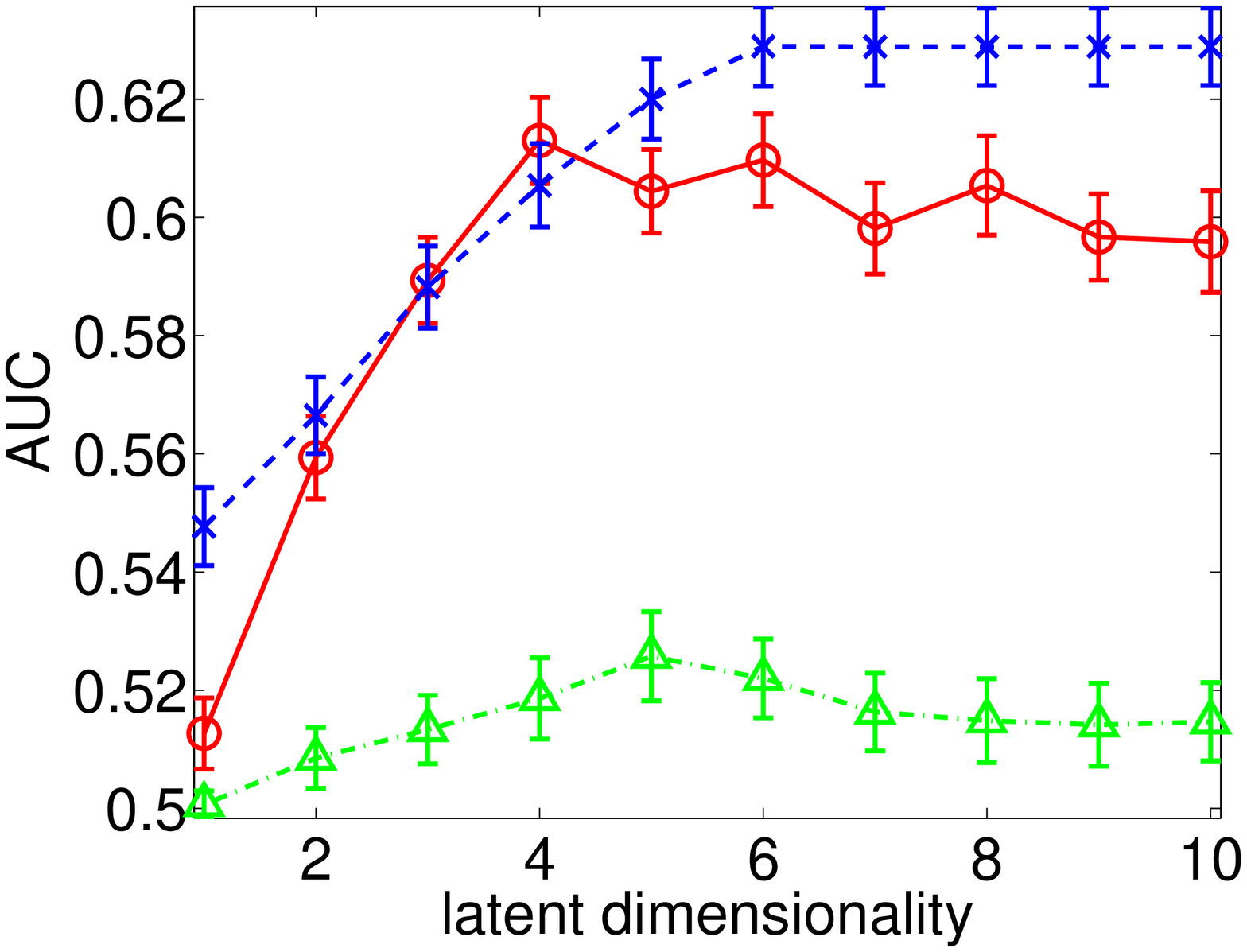} &
\includegraphics[height=8em]{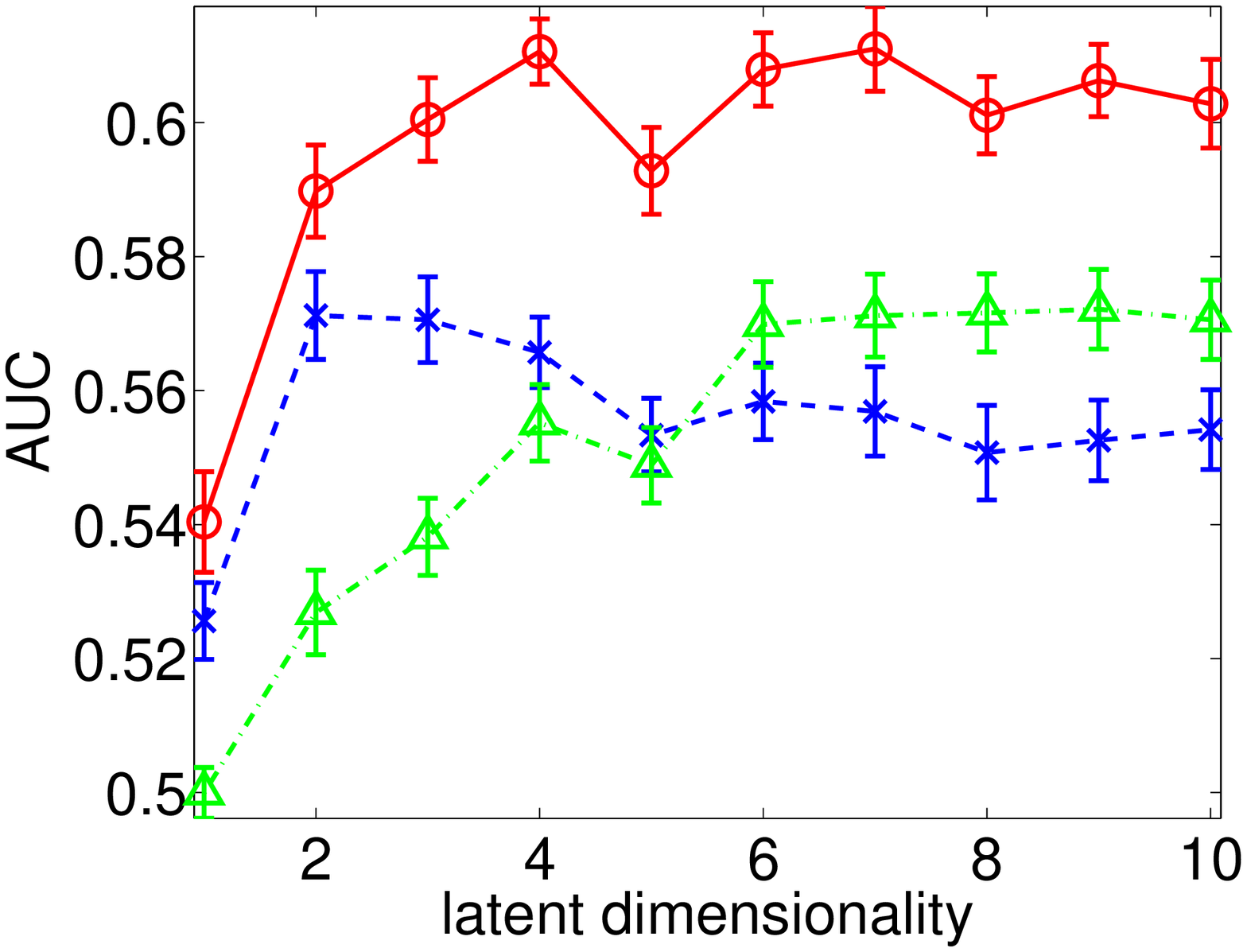} &
\includegraphics[height=8em]{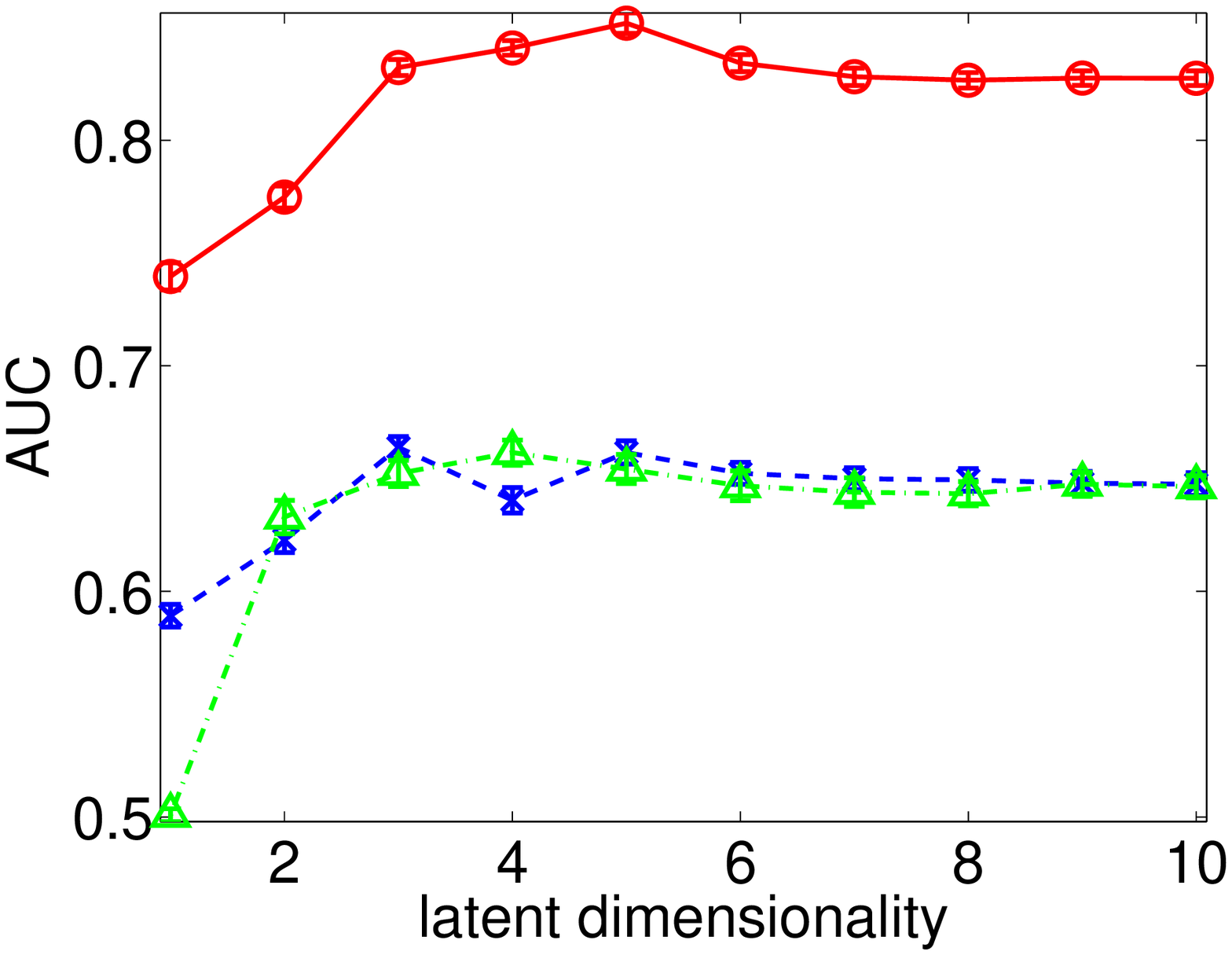} \\
(f) sonar & (g) svmguide2 & (h) svmguide4 \\
\includegraphics[height=8em]{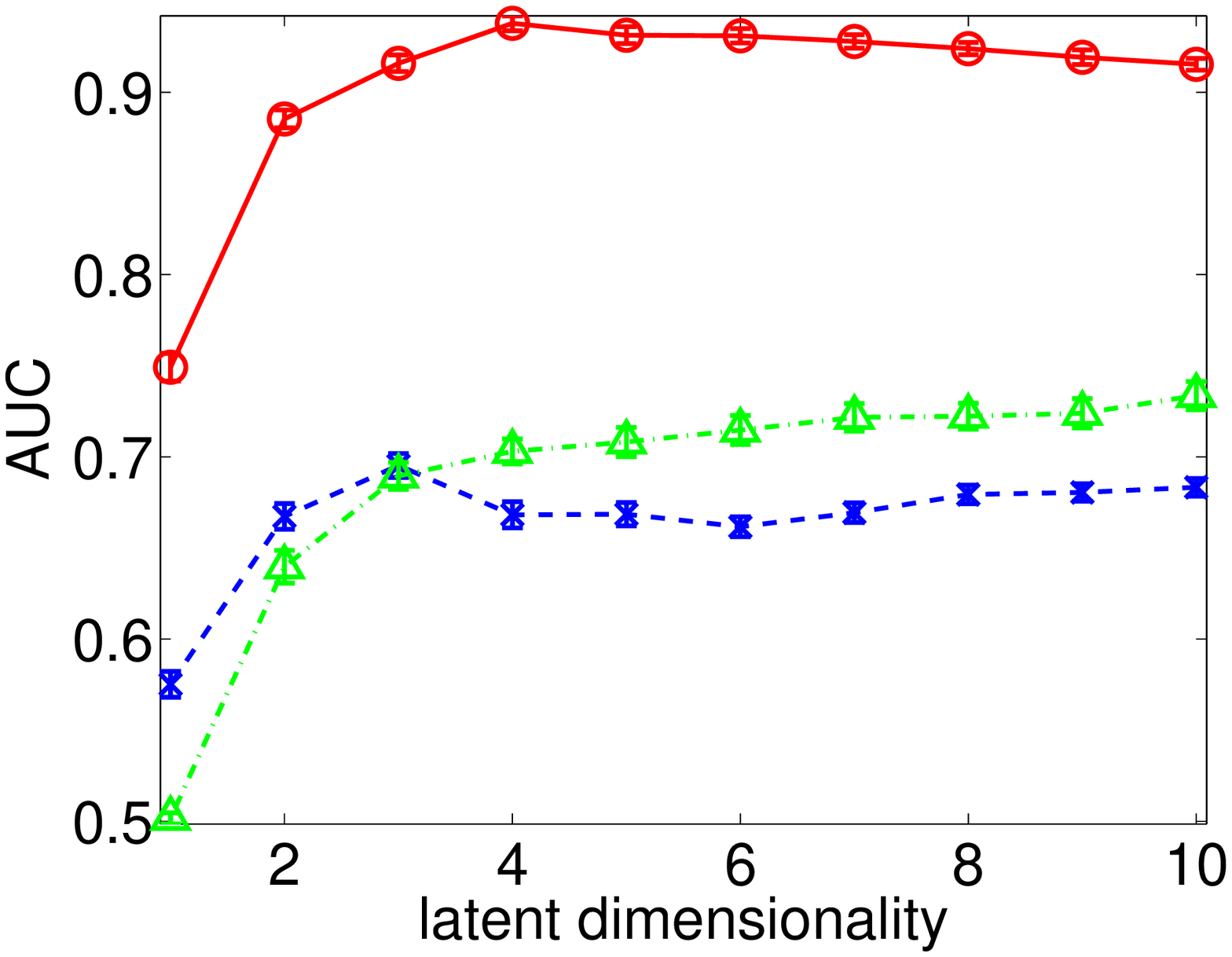} &
\includegraphics[height=8em]{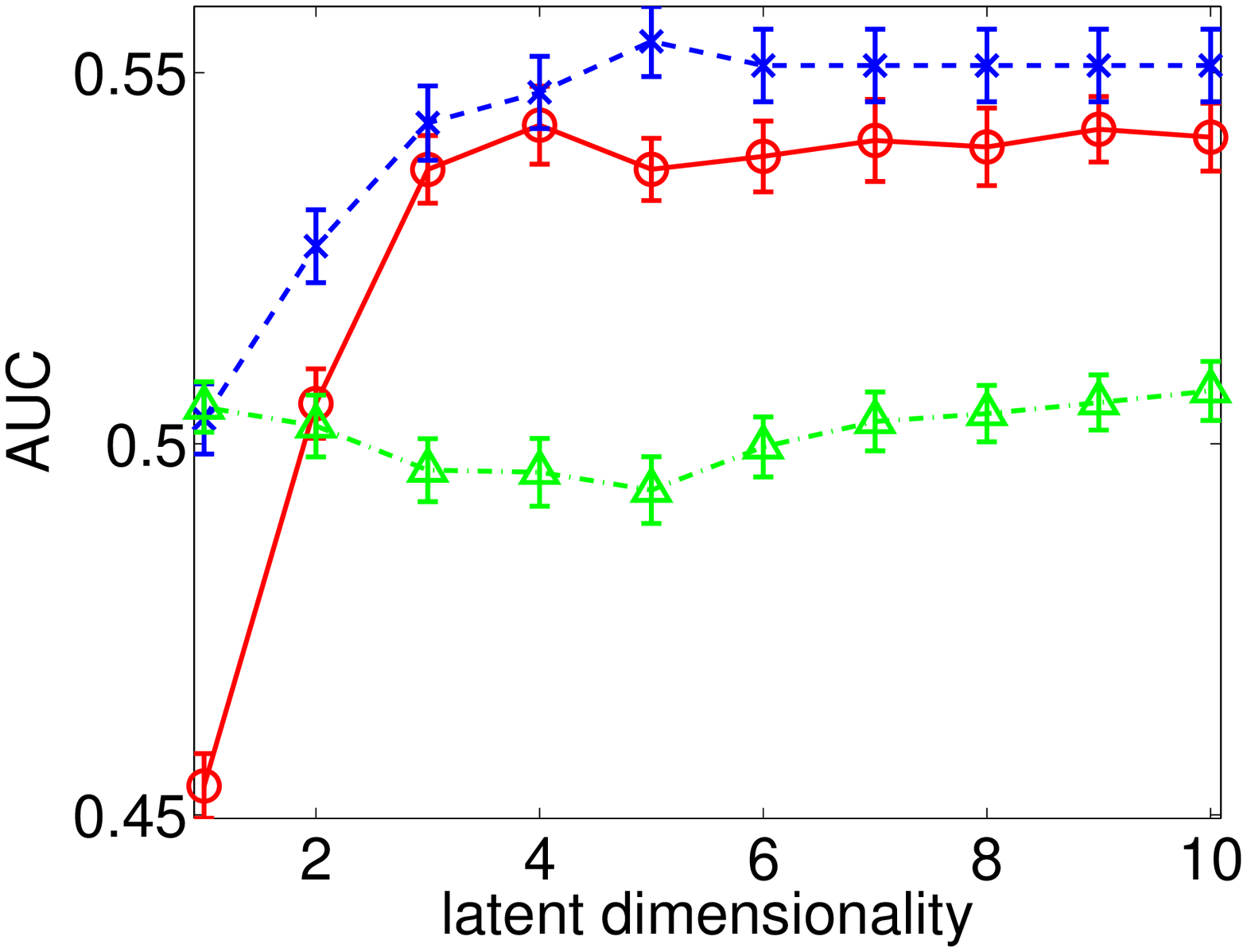} &
\includegraphics[height=8em]{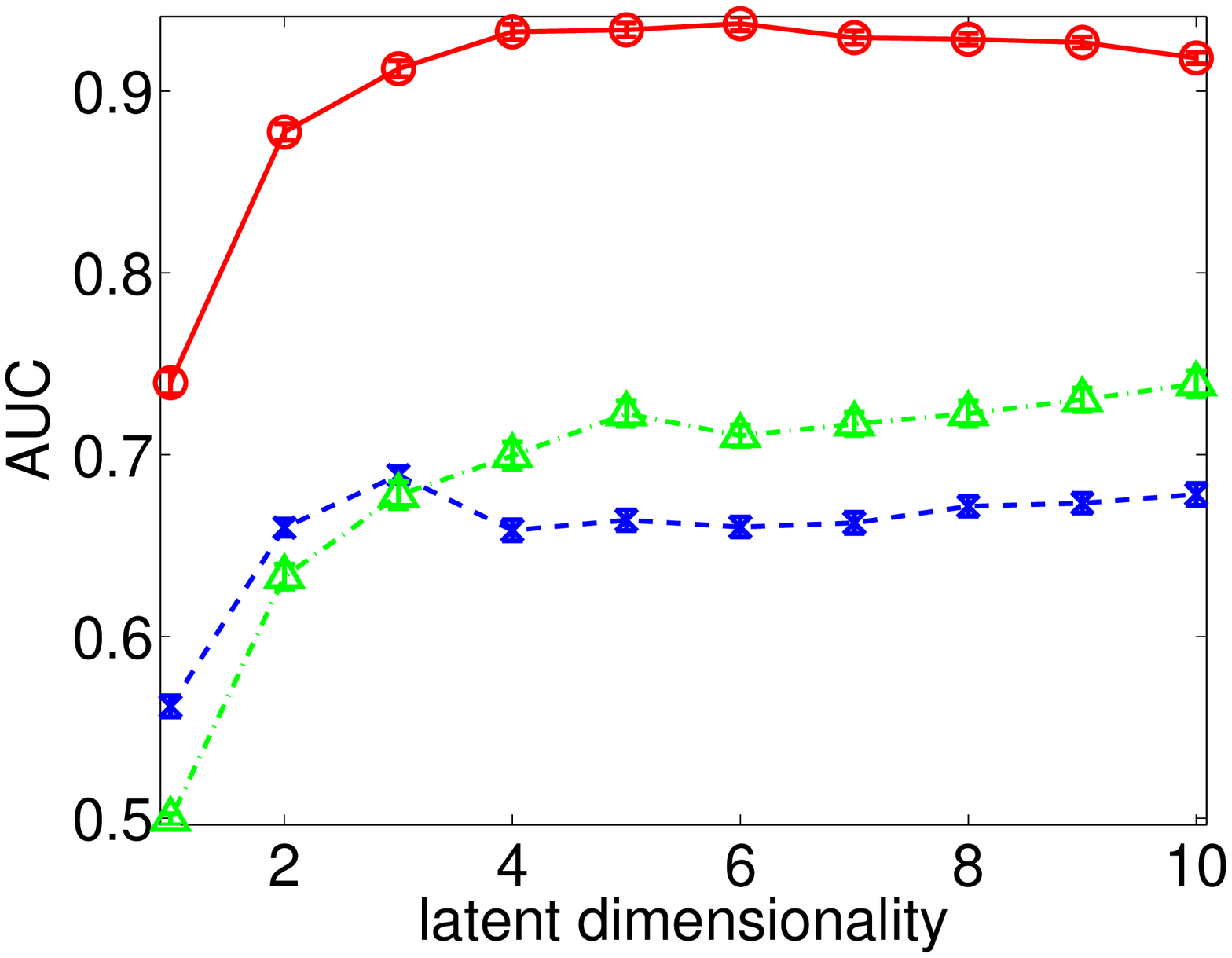} \\
(i) vehicle & (j) vowel & (k) wine \\
\includegraphics[height=8em]{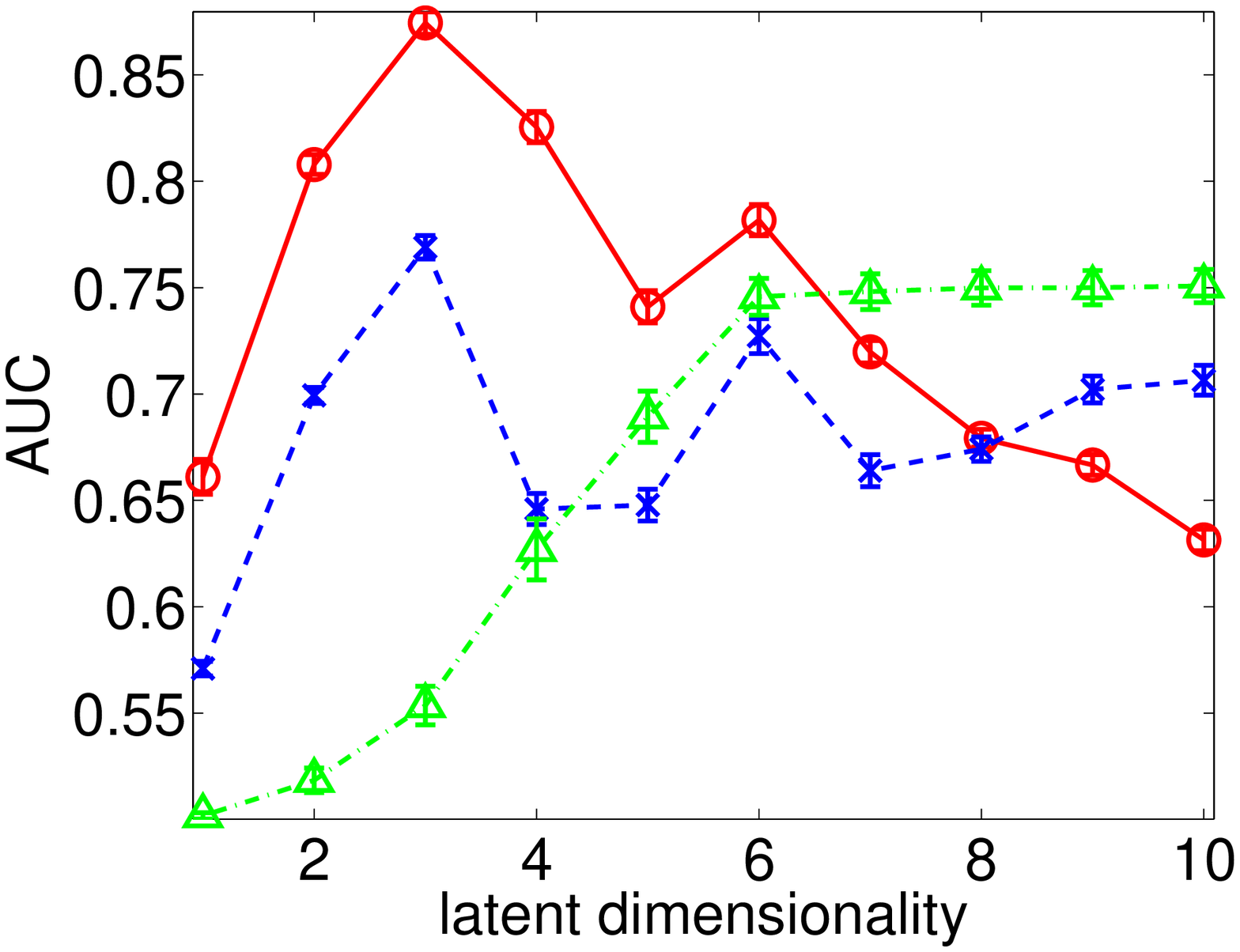} &
\includegraphics[height=8em]{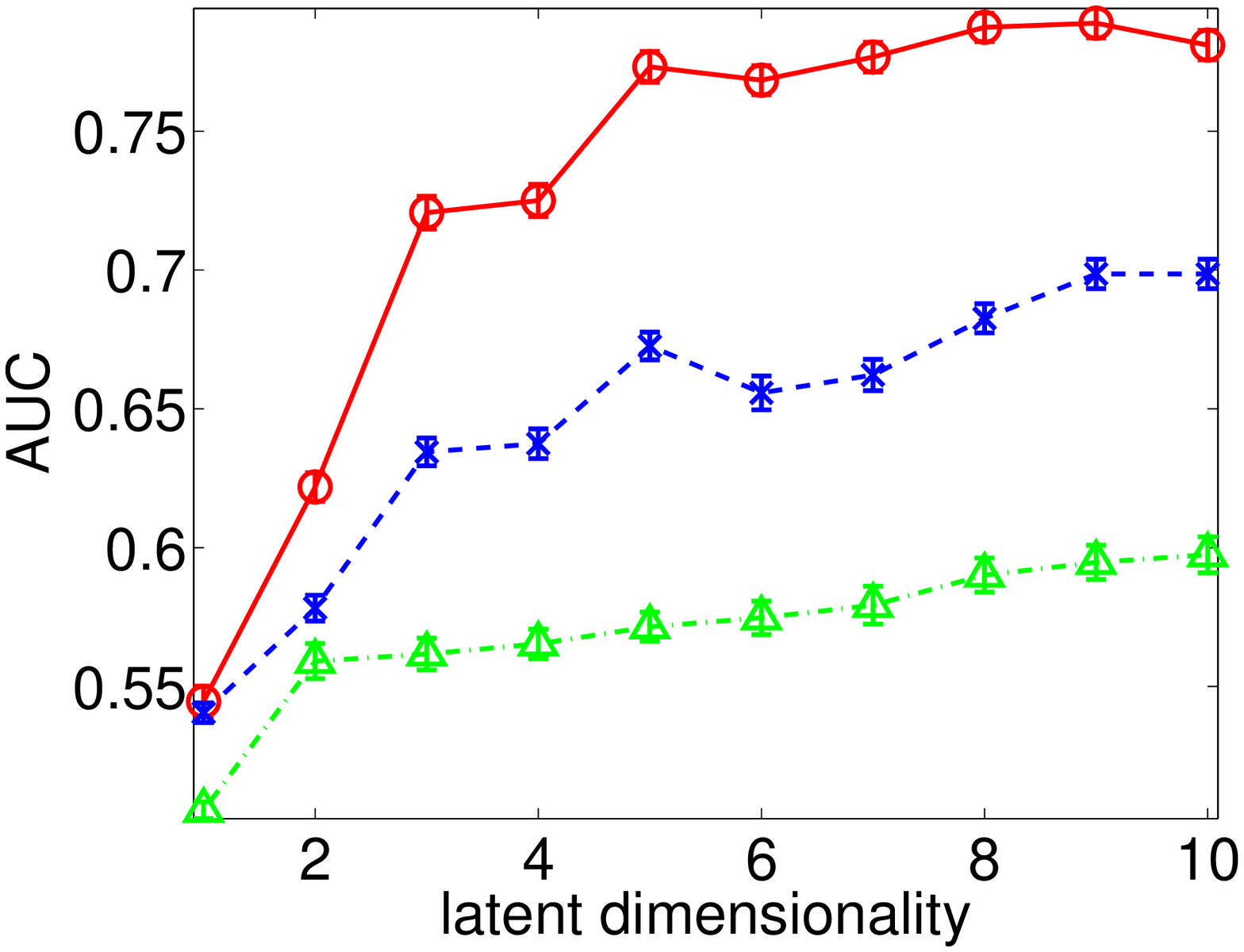} &
\includegraphics[height=8em]{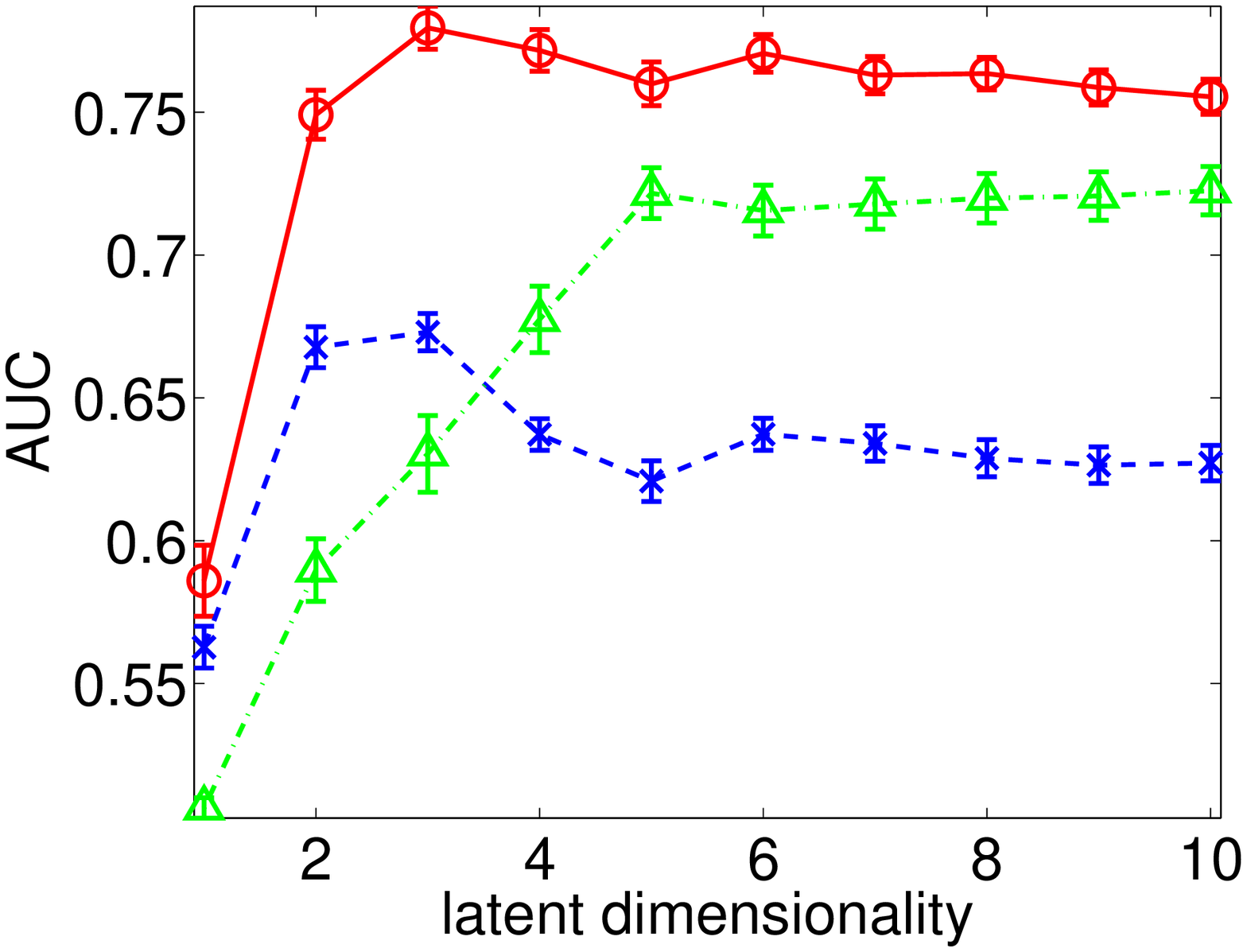} \\ 
\end{tabular}}
\caption{Average AUCs with different dimensionalities of latent vectors, and ther standard errors.}
\label{fig:auc_K}
\end{figure*}

Figure~\ref{fig:auc_K} shows AUCs with different dimensionalities
of latent vectors using data sets whose anomaly rate is 0.4.
When the dimensionality was very low ($K=1$ or $2$),
the AUC was low in most of the data sets, 
because low-dimensional latent vectors cannot represent
the observation vectors well.
With all the methods, the AUCs were relatively stable 
when the latent dimensionality was higher than four.

\subsection{Single-view anomaly detection}
We illustrated that the proposed model does not detect single-view anomalies
using synthetic single-view anomaly data.
With the synthetic data, 
latent vectors for single-view anomalies 
were generated from ${\cal N}(\bm{0},\sqrt{10}\bm{I})$,
and those for non-anomalous instances were generated from 
${\cal N}(\bm{0},\bm{I})$.
Since each of the anomalies has only one single latent vector,
it is not a multi-view anomaly.
The numbers of anomalous and non-anomalous instances were 5 and 95, 
respectively. The dimensionalities of the observed and latent spaces
were five and three, respectively.
Table~\ref{tab:single_auc} shows the average AUCs with the single-view anomaly data, which are averaged over 50 different data sets.
The low AUC of the proposed model indicates that it does not consider
single-view anomalies as anomalies.
On the other hand, the AUC of the one-class SVM (OCSVM) was high
because OCSVM is a single-view anomaly detection method,
and it leads to low multi-view anomaly detection performance.

\begin{table}[t!]
\caption{Average AUCs for single-view anomaly detection.}
\label{tab:single_auc}
\centering
\begin{tabular}{rrr}
\hline
Proposed & PCCA & OCSVM \\
\hline
$\mathbf{0.117} \pm 0.098$ &
$0.174 \pm 0.095$ &
$0.860 \pm 0.232$ \\
\hline
\end{tabular}
\end{table}

\subsection{Missing value imputation}
We can use the proposed model for missing value imputation
in multi-view data with anomalies.
The proposed model was evaluated by comparison with PCCA and Average,
which estimates missing values by averaging the observation 
features over instances.
We randomly selected 5\% of the values as missing.
Table~\ref{tab:mse0.4} shows the mean squared errors 
for missing value imputation with data sets whose anomaly rate is 0.4.
For all of the data sets, the proposed model achieved the lowest error.
This result shows the effectiveness of the proposed model
for imputing missing values in multi-view data with anomalies.

\begin{table*}[t!]
\caption{Average mean squared errors for imputing missing values. Values in bold typeface are statistically better at the $5\%$ level from those in normal typeface as indicated by a paired t-test.}
\label{tab:mse0.4}
\centering
{\tabcolsep=0.4em
\begin{tabular}{l r r r r r r}
\hline
& breast-cancer  & diabetes  & glass  & heart  & ionosphere  & sonar \\
\hline
Proposed & $\mathbf{0.532}$ & $\mathbf{0.052}$ & $\mathbf{0.073}$ & $\mathbf{0.797}$ & $\mathbf{0.771}$ & $\mathbf{1.510}$ \\
PCCA & $0.574$ & $0.063$ & $0.096$ & $1.146$ & $0.935$ & $1.800$ \\
Average & $2.190$ & $0.313$ & $1.045$ & $3.305$ & $2.708$ & $3.754$ \\
\hline
& svmguide2  & svmguide4  & vehicle  & vowel  & wine & \\
\hline
Proposed & $\mathbf{0.003}$ & $\mathbf{1.497}$ & $\mathbf{0.118}$ & $\mathbf{0.150}$ & $\mathbf{0.201}$ & \\
PCCA & $\mathbf{0.003}$ & $1.680$ & $\mathbf{0.177}$ & $0.208$ & $0.255$ & \\
Average & $0.004$ & $3.717$ & $1.767$ & $0.521$ & $1.419$ & \\
\hline
\end{tabular}}
\end{table*}

\subsection{Latent dimensionality estimation}
\begin{figure*}[t!]
\centering
\begin{tabular}{cc}
(a) anomaly rate: 0.2 & (b) anomaly rate: 0.4\\
\includegraphics[height=8em]{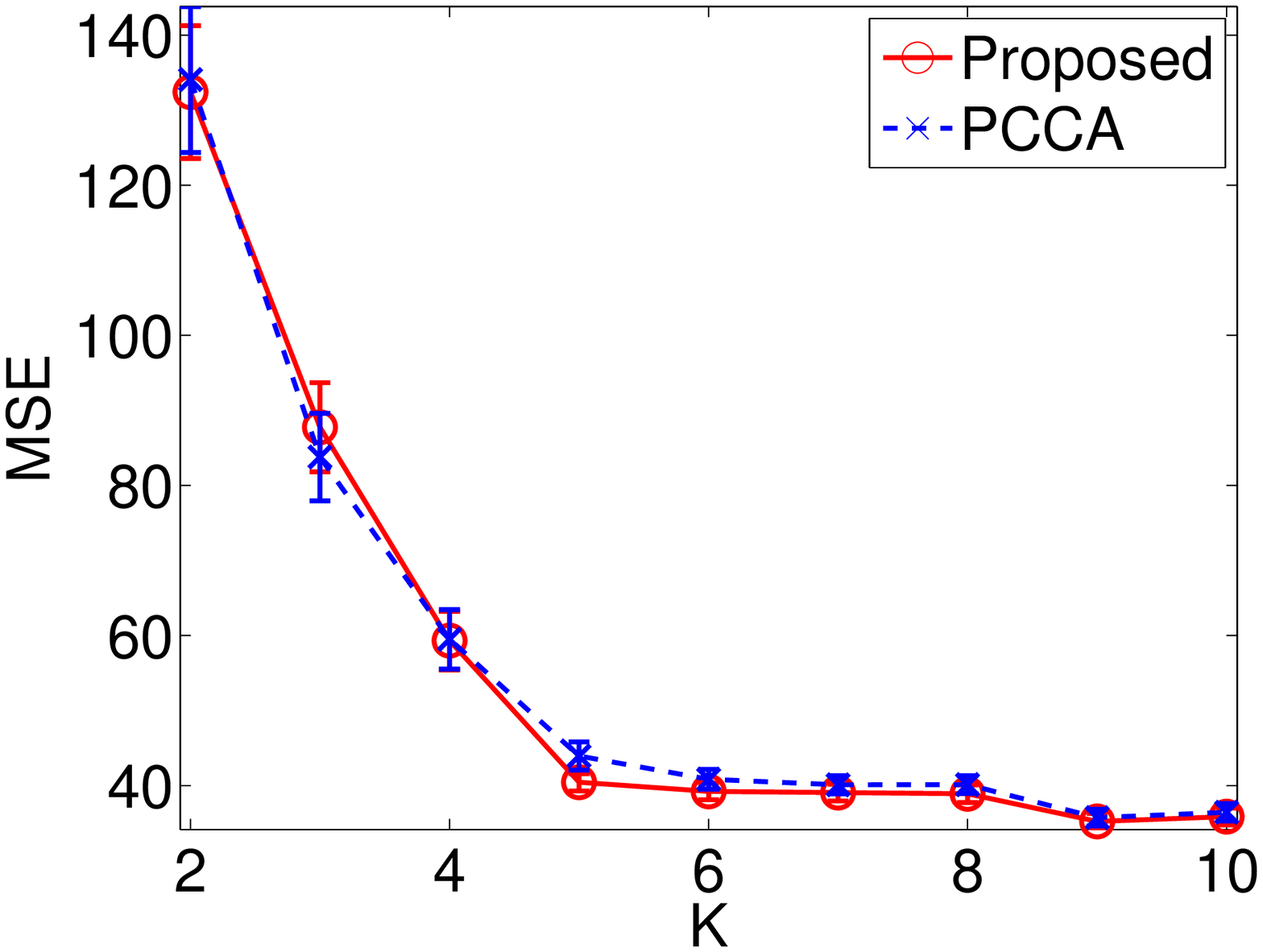} &
\includegraphics[height=8em]{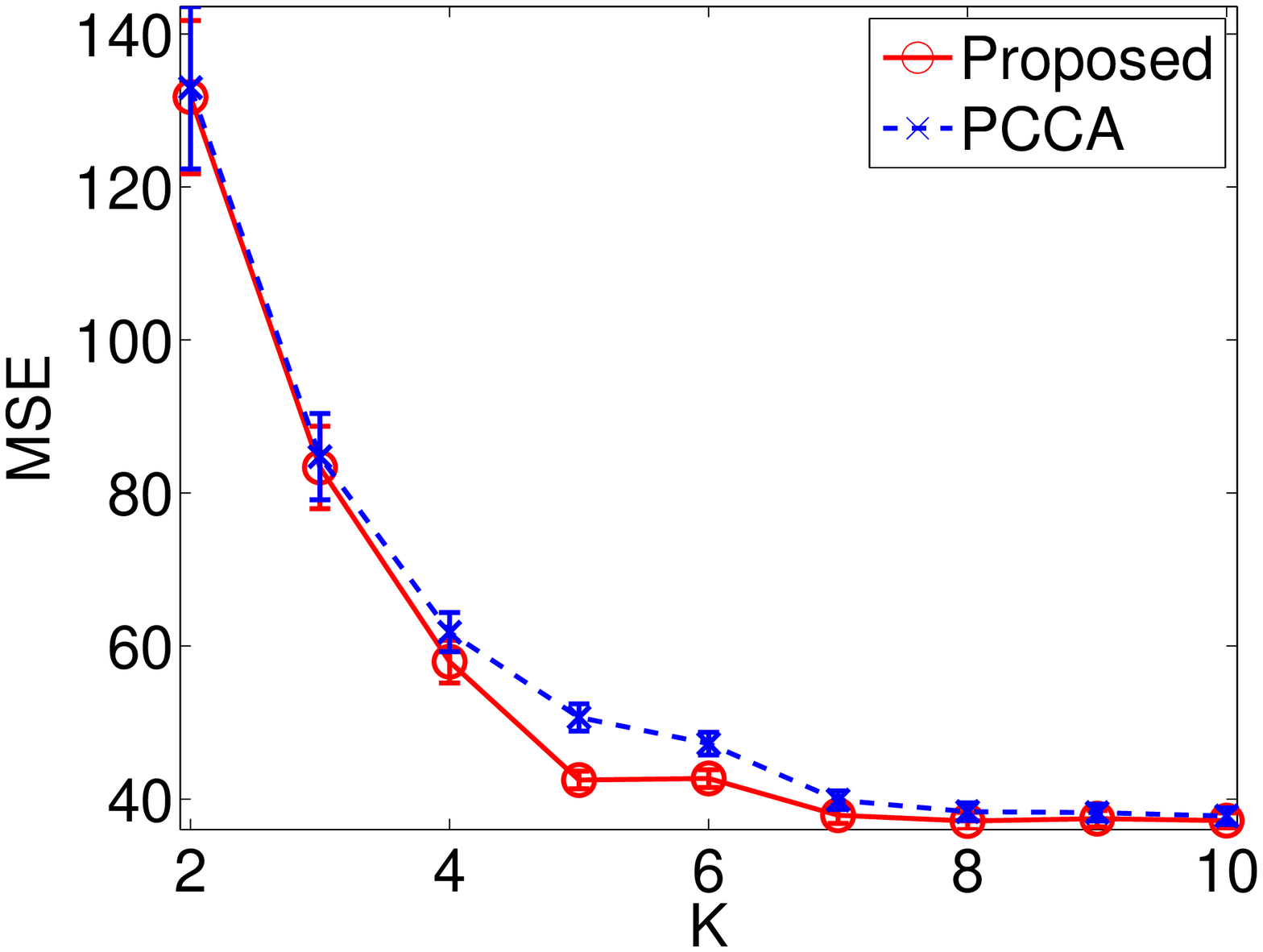}
\\
(c) anomaly rate: 0.6 & (d) anomaly rate: 0.8 \\
\includegraphics[height=8em]{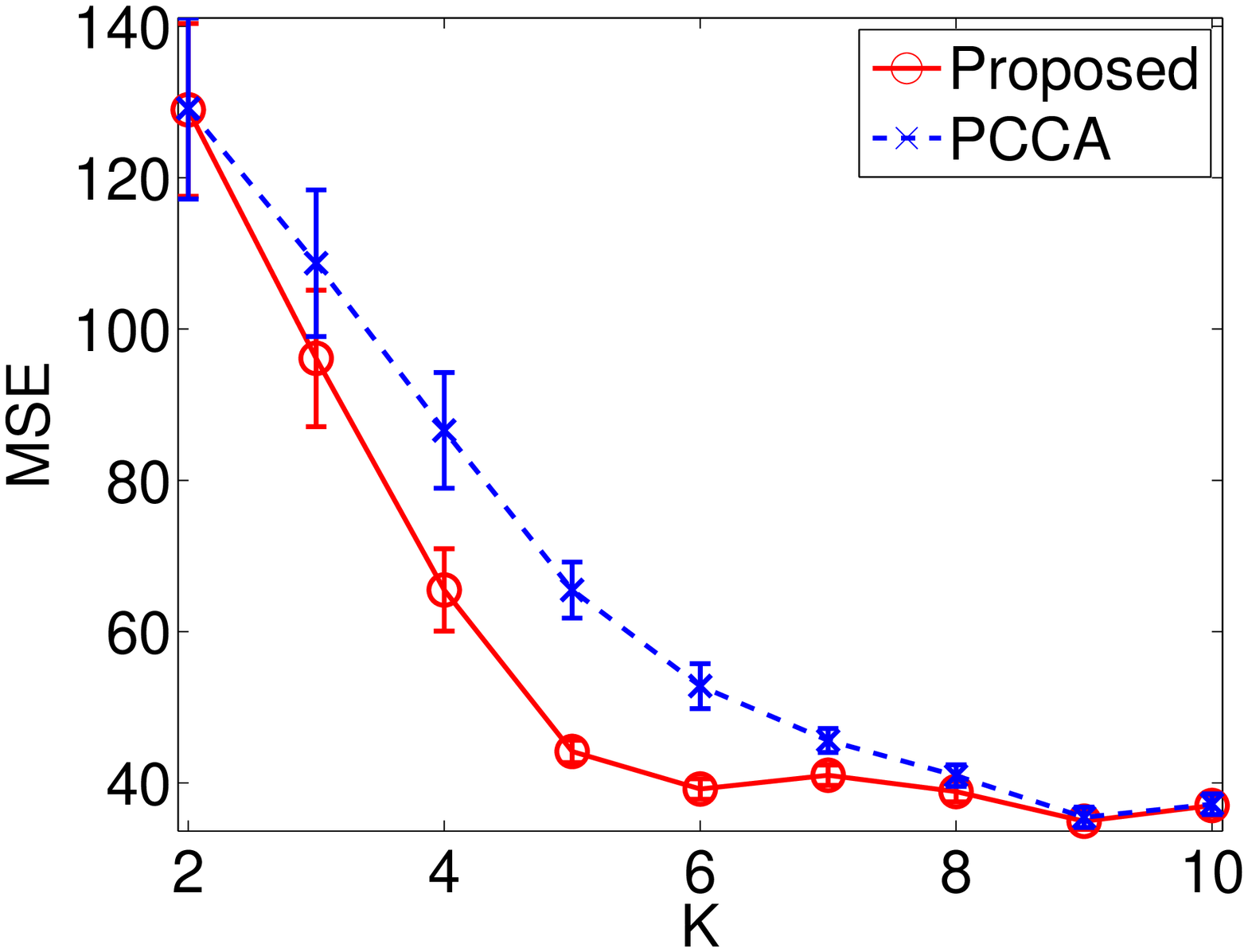} &
\includegraphics[height=8em]{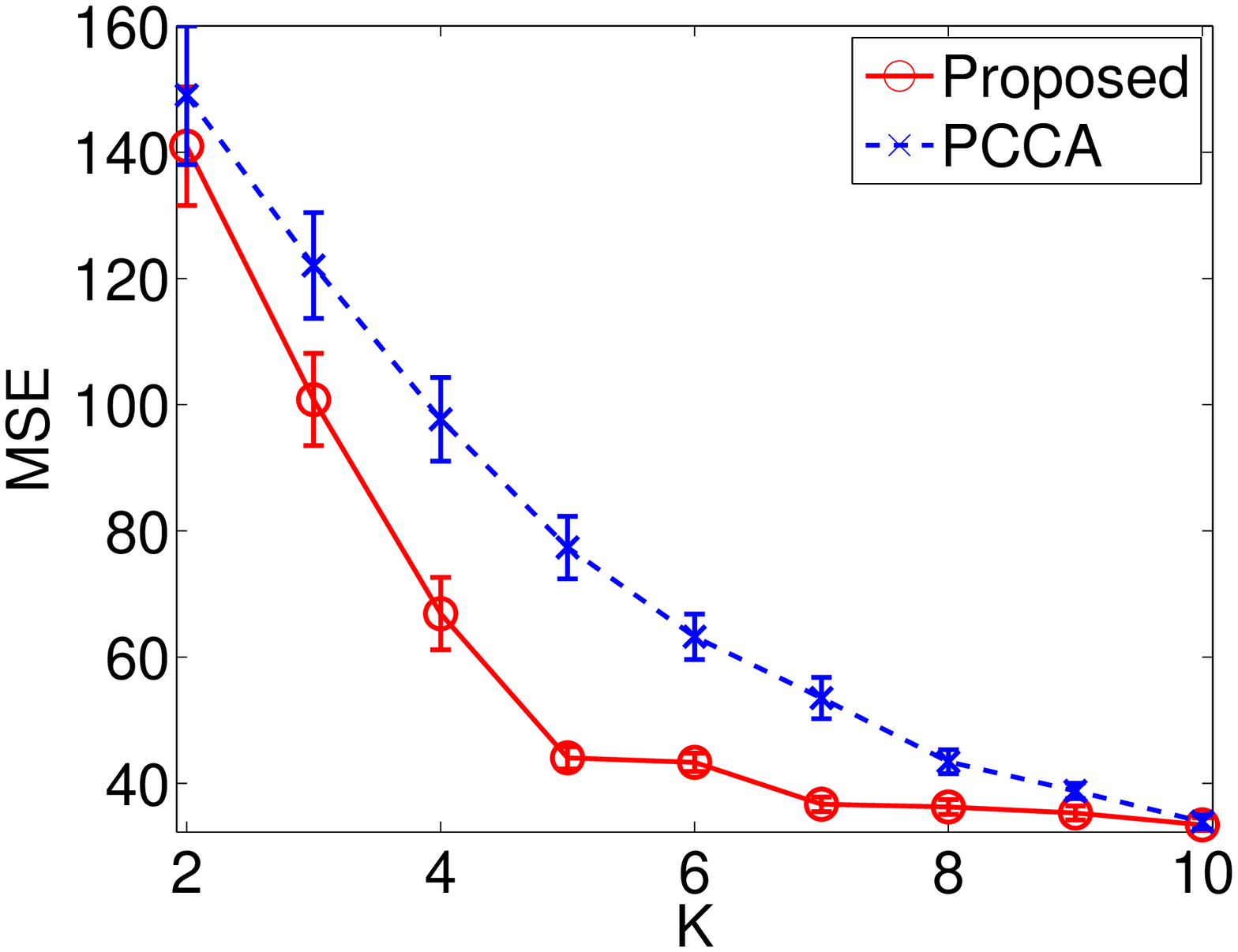}
\\
\end{tabular}
\caption{Average mean squared errors for imputing missing values and their standard errors with different latent dimensionalities using synthetic data sets with different anomaly rates.}
\label{fig:mse_K}
\end{figure*}

We also evaluated the performance of the proposed model
to estimate the latent dimensionality from the given data.
For this purpose,
we used synthetic data sets, 
where the true latent dimensionality was known.
Synthetic data sets with two views were generated 
according to the following procedure.
First, we sampled a latent vector for each non-anomalous instance,
and two latent vectors for each anomaly 
from a $K^{\star}$-dimensional Gaussian distribution with mean $\bm{0}$
and covariance $\bm{I}$.
Here, $K^{\star}$ is the true latent dimensionality,
and we used $K^{\star}=5$.
Second, we generated projection matrices $\bm{W}$
where each element was drawn from a Gaussian distribution with mean $0$
and variance 1. 
Finally, we generated observations 
using the latent vectors and projection matrices with Gaussian noise.
Figure~\ref{fig:mse_K} shows the mean squared errors
for missing value imputation 
using synthetic data sets with different anomaly rates
when the latent dimensionality of the proposed model was changed,
$K\in\{2,\cdots,10\}$.
With the proposed model,
the mean squared errors decreased rapidly when $K<5$,
and they did not vary greatly when $K\geq5$,
in all of the data sets with different anomaly rates.
This result indicates that the proposed model can estimate the 
latent dimensionality from multi-view data with anomalies.
In contrast, with PCCA, 
the mean squared errors decreased even when $K>5$
especially in data sets with high anomaly rates.
This implies that 
PCCA requires a higher latent dimensionality for modeling data with anomalies
than the true latent dimensionality.
When the anomaly rate is small, 
the proposed model and PCCA perform similarly as shown in Figure 6 (a).
Because we use Bayesian inference for the proposed model and PCCA 
based on Markov chain Monte Carlo in our experiments,
we can avoid overfitting when the latent dimensionality is higher
than the true dimensionality
as shown in Bayesian probabilistic matrix factorization~\citep{salak2008}.

\subsection{Application to movie data}
For an application of multi-view anomaly detection,
we analyzed inconsistency between movie rating behavior and genre
in MovieLens data~\citep{herlocker1999algorithmic}.
An instance corresponds to a movie,
where the first view represents whether the movie is rated or not by users,
and the second view represents the movie genre.
Both views consists of binary features.
We used 338 movies, 943 users and 19 genres.
Table~\ref{tab:movie} shows high and low anomaly score movies
when we analyzed the movie data by the proposed method with $K=5$.
`The Full Monty' and `Liar Liar' were categorized in `Comedy' genre.
They are rated by not only users who likes `Comedy',
but also who likes `Romance' and `Action-Thriller'.
`The Professional' was anomaly because 
it was rated by two different user groups, where a group prefers `Romance' 
and the other prefers `Action'.
Since `Star Trek' series are typical Sci-Fi and liked by specific users,
its anomaly score was low.

\begin{table}[t]
\centering
\caption{High and low anomaly score movies calculated by the proposed model.}
\label{tab:movie}
{\tabcolsep=0.4em
 \begin{tabular}{lr|lr}
\hline
 Title & Score & Title & Score \\
\hline
The Full Monty & 0.98 & Star Trek VI & 0.04 \\
Liar Liar & 0.93 & Star Trek III & 0.04 \\
The Professional & 0.91 & The Saint & 0.04 \\
Mr. Holland's Opus & 0.88 & Heat & 0.03 \\
Contact & 0.87 & Conspiracy Theory & 0.03 \\
\hline
 \end{tabular}}
\end{table}

\section{Conclusion}
\label{sec:conclusion}

We proposed a generative model approach for multi-view anomaly detection,
which finds instances that have inconsistent views.
In the experiments,
we confirmed that the proposed model could perform much better than
existing methods for detecting multi-view anomalies,
and for imputing missing values in multi-view data with anomalies.
There are several avenues that can be pursued for future work.
The proposed model assumes the linearity of observations with respect 
to their latent vectors. We can relax this assumption by using 
Gaussian processes~\citep{lawrence2004gaussian,rasmussen2005gaussian,shon2006learning}.
We would like to evaluate our proposed model for other real applications.
CCA is successfully used for a wide variety of real multi-view data, 
such as multilingual data and image-annotation data. 
This indicates that the proposed model based on CCA 
can handle the same wide variety of real data.



\bibliographystyle{spbasic}      
\bibliography{ml2014}

\end{document}